\documentclass[10pt,logo,copyright]{nvidiatechreport}
\usepackage[authoryear,round]{natbib}

\usepackage[utf8]{inputenc} 
\usepackage[T1]{fontenc}    

\usepackage{parskip}        
\usepackage{url}            
\usepackage{booktabs}       
\usepackage{amsfonts}       
\usepackage{nicefrac}       
\usepackage{microtype}      
\usepackage{xcolor}         
\usepackage[dvipsnames]{xcolor} 
\usepackage{graphicx}
\usepackage{animate}        
\usepackage{subcaption}
\usepackage{tabularx}
\usepackage{makecell}
\usepackage{adjustbox}
\usepackage{setspace}
\newcolumntype{M}[1]{>{\centering\arraybackslash}m{#1}}
\usepackage{float}
\usepackage{tikz}
\usetikzlibrary{positioning,shapes,arrows}
\usepackage{amsmath,amsfonts,bm, bbm,leftindex}
\usepackage{multirow}
\usepackage{comment}
\usepackage{gensymb}
\usepackage{lipsum}
\usetikzlibrary{arrows.meta, positioning, fit}
\usepackage[para]{threeparttable}
\usepackage{tikz}
\usetikzlibrary{tikzmark}

\usepackage[most]{tcolorbox}
\usepackage{fancyvrb}
\usepackage{fvextra}
\usepackage{dashrule}

\usepackage{fontawesome5}
\usepackage{hyperref}

\usepackage{pifont}
\usepackage{wrapfig}

\usepackage[nameinlink]{cleveref}
\crefname{equation}{Eq.}{Eqs.}
\crefname{figure}{Fig.}{Figs.}
\crefname{section}{Sec.}{Sec.}
\crefname{appendix}{App.}{App.}
\crefname{table}{Tab.}{Tabs.}
\crefname{algorithm}{Algo}{Algo}
\crefname{thm}{Thm}{Thm}
\Crefname{thm}{Thm}{Thm}
\crefname{prop}{Prop}{Prop}
\usepackage[ruled,vlined]{algorithm2e}

\definecolor{darkred}{rgb}{0.7, 0.0, 0.0}

\newcommand{\crefnames}[3]{%
  \@for\next:=#1\do{%
    \expandafter\crefname\expandafter{\next}{#2}{#3}%
  }%
}

\title{ThinkWorld: World Imagination Reinforces Chain-of-Dynamics Reasoning}

\author{
     {Chaehyun Kim,}$^{1,2}$\quad {Chi-Pin Huang,}$^{1}$\quad {Jisoo Kim,}$^{2}$\quad {Min-Hung Chen,}$^{1}$ \par \vspace{-8pt}
     {Yu-Chiang Frank Wang,}$^{1}$\quad {Seungryong Kim,}$^{2}$\quad {Fu-En Yang}$^{1}$ \\
    \small \textsuperscript{1} NVIDIA \quad
    \small \textsuperscript{2} KAIST
}

\correspondingauthor{X}

\begin{document}

\maketitle

\begin{figure*}[h]

    \vspace{-19pt}

    \centering

    \includegraphics[width=\linewidth]{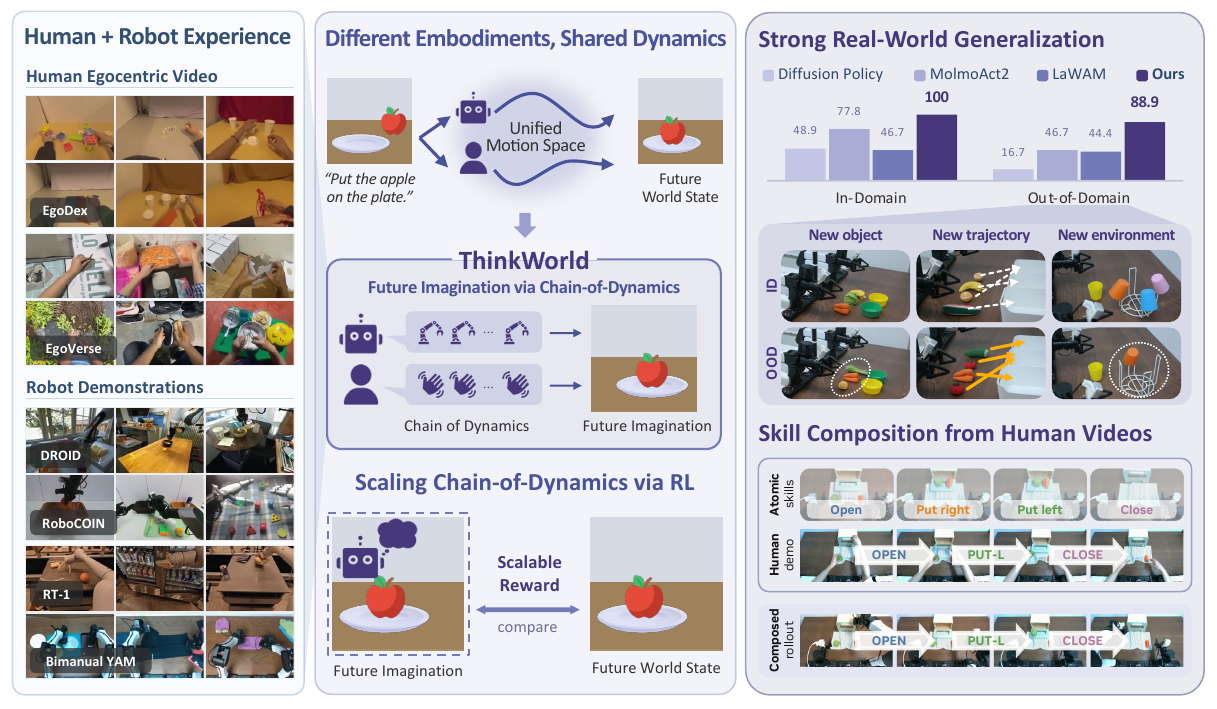}
    \vspace{-22pt}
    \caption{We introduce \textbf{ThinkWorld}, a vision-language-action framework that learns transferable Chain-of-Dynamics (CoD) reasoning from diverse human and robot experiences. CoD captures action-induced dynamics across embodiments and is reinforced through future imagination, enabling robust generalization in the real world and skill composition guided by human videos.}
    \vspace{-13pt}
    \label{fig:teaser}
\end{figure*}

\begin{abstract}
Vision-Language-Action (VLA) models have made substantial progress in robotic manipulation, but their generalization remains constrained by the diversity and coverage of action-labeled robot demonstrations. Human videos provide much broader behavioral and environmental experience, while learning from them faces a fundamental trade-off between scalable data and explicit action grounding. We introduce \emph{ThinkWorld}, a VLA learning framework that grounds action reasoning in world dynamics and strengthens it through reinforcement learning with world imagination. ThinkWorld represents each action thought as a Chain-of-Dynamics (CoD), a sequence of discrete motion tokens that captures the intermediate dynamics connecting the current world to a future state and serves as a common action representation across embodiments. We first initialize CoD with action-labeled demonstrations, then predict the future induced by each CoD and reinforce those whose predictions better match the observed outcome. This enables videos without action annotations to provide direct supervision from observed state transitions, allowing human and robot experience to refine the same reasoning space. Experiments in simulation and on real robots demonstrate effective human-to-robot transfer and strong generalization to unseen tasks and environments.\\

\vspace{-10pt}
\textbf{Links:} \hspace{2pt}
{
\hypersetup{urlcolor=nvidiagreen}
\href{https://thinkworld-project.github.io/}{Project Page}
}

\end{abstract}
\abscontent
\section{Introduction}

Building general-purpose action models that transfer across tasks, objects, environments, and embodiments is a central goal of robot learning. Recent vision-language-action models (VLAs)~\citep{brohan2023rt,o2024open,team2024octo,kim24openvla,black2024pi_0,intelligence2025pi0,bjorck2025gr00t} have made substantial progress by transferring pretrained vision-language knowledge and scaling across heterogeneous robot demonstrations. Nevertheless, their policy learning and behavioral coverage remain largely tied to action-labeled robot trajectories, whose collection requires physical platforms, embodiment-specific interfaces, and substantial human effort. Human videos offer a complementary source of experience with far greater behavioral and environmental diversity, but do not provide directly executable robot actions.

Existing approaches seek to extract actionable supervision from human videos along two main directions. Latent action models~\citep{ye2025latent,bu2025univla,dai2026conla,sun2026vla} infer transition variables from pairs or sequences of frames and use them for representation or policy learning. While naturally scalable to action-unlabeled videos, these inferred variables are underconstrained and may capture appearance changes or camera motion rather than controllable behavior. In contrast, fully supervised approaches recover hand poses, object interactions, or retargeted trajectories to provide explicit motion supervision~\citep{kareer2025egomimic,zheng2026egoscale,liu2026egoengine}. Such supervision provides direct physical grounding, but relies on accurate pose estimation and human-to-robot correspondence. Human-video learning therefore yields a trade-off between stronger action grounding from explicit supervision and broader behavioral coverage from unlabeled videos.

Future observations provide a natural supervisory signal to bridge this gap as every video records the consequences that follow a behavior, regardless of embodiment. Predictive world models~\citep{agarwal2026cosmos,zhou2024dino,assran2025vjepa2} and foresight-based VLAs~\citep{zhang2026dreamvla,sun2026vla,cai2026internvla,li2026fomovla} exploit this signal by forecasting future pixels, visual features, or trajectories, providing a direct account of \emph{what changes} in the world. However, inferring the intermediate dynamics that result in these changes remains challenging, particularly from large-scale human videos without action annotations. Recent reasoning-augmented VLAs offer a useful perspective by introducing intermediate tokens, such as textual plans~\citep{zawalski2024robotic,huang2026thinkact}, visual guidance~\citep{lee2025molmoact,fang2026molmoact2}, continuous latents~\citep{huang2026fast}, or coarse action trajectories~\citep{zhong2026acot}, to structure action generation. This raises a natural question of whether such intermediate representations can instead be grounded in the dynamics they induce in the world, providing a shared action interface across embodiments.

Motivated by this perspective, we introduce \emph{ThinkWorld}, a VLA learning framework that formulates action thoughts as Chain-of-Dynamics (CoD) and reinforces them through world imagination. CoD represents an action as a sequence of discrete motion tokens that captures how the world evolves over time, providing a common reasoning interface across human and robot behaviors. We first initialize this space with action-labeled demonstrations, grounding CoD in physically meaningful behavior. ThinkWorld then predicts the future world representation induced by each CoD and compares it with the future observation, using their alignment to reinforce action reasoning. As a result, videos even without motion annotations provide a learning signal, allowing large-scale human videos and robot trajectories to jointly refine the same CoD space. This training objective propagates through the shared VLM, enriching the representations for action prediction with knowledge of how motion transforms the world. Across simulation benchmarks and real-robot evaluations, we demonstrate that ThinkWorld enables human-to-robot skill composition and generalization to unseen scenarios.

Our contributions are as follows:
\begin{itemize}
    \item We introduce \emph{ThinkWorld}, a VLA learning framework that couples a common VLM-native action-reasoning interface for both human and robot with internal world imagination.
    \item We propose consequence-grounded reinforcement learning, which rewards CoD according to its consistency with observed future consequences and thereby enables action reasoning to improve from videos without action labels.
    \item We show that ThinkWorld benefits from scaling action-unlabeled human video and enables human-to-robot skill transfer, allowing robots to execute complete tasks demonstrated in human videos while receiving robot supervision only for the constituent atomic skills.
\end{itemize}

\section{Related Works}
\label{sec:related_works}

\textbf{Generalist Robot Policies.}
Vision-Language-Action (VLA) models~\citep{kim24openvla,bjorck2025gr00t,black2024pi_0,intelligence2025pi0,wang2026qwen,chen2025internvla,team2025gemini} build upon advances in vision-language models (VLMs)~\citep{bai2025qwen3,li2025eagle,chen2026eagle,liu2025nvila,ye2026omnivinci} to learn generalist manipulation policies that map visual observations and language instructions to executable actions..
Despite acquiring broad capabilities from large-scale robot demonstrations, end-to-end imitation remains bounded by the behaviors observed during training.
Recent embodied reasoning methods introduce intermediate reasoning between perception and control~\citep{zawalski2024robotic,huang2026thinkact,lee2025molmoact,huang2026fast,fang2026molmoact2,zhong2026acot,lu2026tacitvla,liu2026long}.
ThinkAct~\citep{huang2026thinkact} reinforces textual and visual plans with action-aligned rewards, MolmoAct~\citep{lee2025molmoact} reasons through spatial traces, Fast-ThinkAct~\citep{huang2026fast} compresses reasoning into latent thoughts, and ACoT-VLA~\citep{zhong2026acot} predicts coarse reference actions to guide policy generation.
However, these methods largely augment robot demonstrations with intermediate reasoning, without explicitly grounding the learned thoughts in the task-relevant changes they induce in the world.
Their generalization to novel scenes and skills therefore remains constrained by the behavioral coverage of the underlying robot data.

\textbf{Learning Robot Policies from Human Videos.}
Improving generalization also requires expanding the behavioral coverage of the training data.
Human videos capture substantially greater diversity in scenes, objects, and skills than robot demonstrations, motivating their use for scaling robot learning~\citep{yang2025egovla,zheng2026egoscale,kim2026ego,li2026egowam}.
Supervised methods convert human interactions into robot-compatible representations, as in EgoScale~\citep{zheng2026egoscale} and EgoWAM~\citep{li2026egowam}, but require costly cross-embodiment supervision.
Latent action models instead infer action-like variables from visual transitions without explicit control labels~\citep{ye2025latent,chen2025moto,routray2026vipra}, with LAPA~\citep{ye2025latent} learning discrete latent actions before grounding them into robot control.
Such representations may capture appearance or embodiment cues rather than task-relevant dynamics. ThinkWorld grounds shared motion tokens and refines them by matching predicted and observed futures in videos.

\textbf{World Models and Visual Foresight for Embodied Control.}
Recent advances in generative video models~\citep{wan2025wan,agarwal2025cosmos} have enabled their application to a wide range of tasks~\citep{xu2023open,tosi2024diffusion,huang2025videomage,kim2025seg4diff}. World models learn action-conditioned dynamics to support planning and policy optimization. Dreamer-style agents optimize policies through imagined rollouts~\citep{hafner2023mastering,hafner2025training}, while recent large-scale models acquire broad physical priors from diverse video data~\citep{agarwal2026cosmos,gao2026dreamdojo,li2026hydra}. In robotics, visual-foresight methods generate video plans or image subgoals to guide control~\citep{du2023learning,black2024zero,bharadhwaj2024gen2act,zhao2025cot}, while predictive VLAs forecast future pixels or latent features for planning and representation learning~\citep{zhang2026dreamvla,sun2026vla,cai2026internvla,li2026fomovla,kim2026pri4r}. More recent world-action models jointly model future-world evolution and robot actions~\citep{ye2026world,pai2025mimic,cen2025worldvla,kim2026cosmos}. In contrast, ThinkWorld does not use future prediction as the control output or an imagined rollout for policy optimization. Instead, it treats agreement between predicted and observed futures as a training-time signal to reinforce action-grounded Chain-of-Dynamics within the VLM. This directly turns observed world changes, including those in action-unlabeled videos, into supervision for action reasoning, while future prediction is discarded at deployment.
\section{Method}
\label{sec:method}

\begin{figure}[t]
    \centering
    \includegraphics[width=\linewidth]{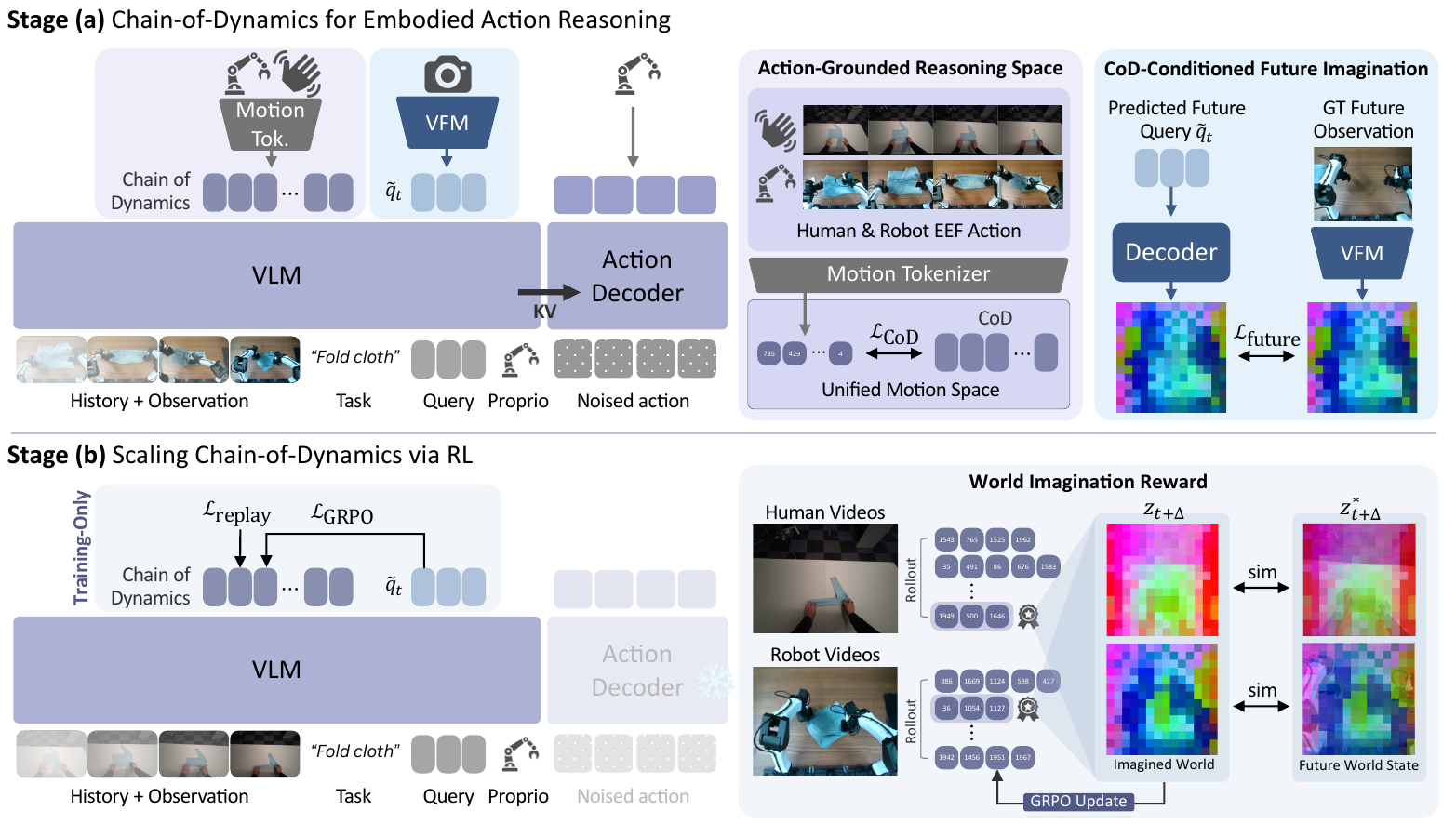}
    \caption{\textbf{ThinkWorld architecture and training stages.} In stage (a), supervised fine-tuning (SFT) tokenizes robot actions and annotated human hand motions into a unified discrete motion space to ground Chain-of-Dynamics (CoD) reasoning, while CoD-conditioned imagination predicts the future visual feature. In stage (b), reinforcement learning (RL) uses the alignment between imagined and observed futures as a scalable reward for CoD reasoning over human and robot videos, strengthening action reasoning without motion annotations and improving guidance for continuous control.}
    \label{fig:architecture}
\end{figure}

\subsection{Problem Formulation}
\label{sec:problem}

Given a language instruction $l$, a current visual observation $o_t$, sampled history frames $\mathbf{h}_t$, and proprioceptive state $s_t$ in case of robot, our goal is to predict a continuous robot action sequence $\mathbf{a}_t$. Rather than treating actions solely as supervision for continuous control, ThinkWorld turns them into a reasoning interface for modeling how an action should unfold and transform the world. Through supervised fine-tuning, we instantiate this interface as a \emph{Chain-of-Dynamics (CoD)} $\mathbf{c}_t$, grounding its discrete motion vocabulary with action-labeled demonstrations across human and robot embodiments while learning CoD-conditioned future imagination through learnable queries that predict the future visual state (Sec.~\ref{sec:cod}). We then reinforce CoD reasoning through world imagination, using the alignment between the imagined future and the observed future $o_{t+\Delta}$ after a physical horizon $\Delta$ as a learning signal that does not require motion annotations (Sec.~\ref{sec:world_rl}). Together, action grounding and reinforcement through world imagination refine the model representations used for continuous control while extending learning to action-unlabeled videos at scale.

\subsection{Chain-of-Dynamics for Embodied Action Reasoning}
\label{sec:cod}

Robot actions are typically represented in embodiment-specific control spaces, such as joint positions or end-effector poses. Although directly executable, these control targets keep action supervision tied to individual embodiments, limiting transfer across robots and from human demonstrations. We therefore introduce \emph{Chain-of-Dynamics (CoD)}, a sequence of discrete motion tokens that provides a shared action language for human and robot motion, as illustrated in Fig.~\ref{fig:architecture}(a).

\textbf{Action-Grounded Reasoning Space.}
Let $\mathbf{m}_t$ denote a motion segment from a robot trajectory or an annotated human hand trajectory. We map both to a normalized delta end-effector representation and tokenize them with a FAST-style motion tokenizer $\mathcal{T}(\cdot)$~\citep{pertsch2025fast}, obtaining the target CoD sequence $\mathbf{c}_t^\star=\mathcal{T}(\mathbf{m}_t)$ in a shared discrete vocabulary. Conditioned on the instruction, visual history, and current observation, the VLM learns to generate $\mathbf{c}_t$ autoregressively. This tokenization provides a common interface grounded in motion from both sources without assuming kinematic equivalence between humans and robots.

\textbf{CoD-Conditioned Future Imagination.}
Motion grounding defines a CoD sequence, but not the world change it induces. To predict this change, we append learnable future queries $\mathbf{q}_{t+\Delta}$ after the CoD. Given the visual context and teacher-forced CoD $\mathbf{c}_t^\star$, the VLM contextualizes these queries as $\tilde{\mathbf{q}}_{t+\Delta}$, which a lightweight decoder $\mathcal{D}_{\mathrm{future}}$ maps to the predicted future representation $\mathbf{z}_{t+\Delta}$ at physical horizon $\Delta$:
\begin{equation}
\begin{aligned}
\tilde{\mathbf{q}}_{t+\Delta} 
&= \text{VLM}(l,\mathbf{h}_t,o_t, \mathbf{c}_t^\star, \mathbf{q}_{t+\Delta}), \\
\mathbf{z}_{t+\Delta}
&= \mathcal{D}_\text{future}(\tilde{\mathbf{q}}_{t+\Delta}).
\end{aligned}
\label{eq}
\end{equation}
Instead of reconstructing future RGB pixels, we predict in a frozen DINO feature space. DINO features retain dense spatial and object structure while abstracting away appearance details~\citep{zhou2024dino,simeoni2025dinov3}, allowing imagined and observed futures to be compared by structural change without photorealistic generation. The target $\mathbf{z}_{t+\Delta}^\star$ is obtained by encoding the observed future frame $o_{t+\Delta}$ with the frozen DINO encoder $\mathcal{E}_{\mathrm{future}}$. This frame is used only to construct the target and is never provided as VLM input.

\textbf{Joint Supervision of CoD and Future Prediction.}
We jointly train the VLM to generate CoDs from observed motion and predict future visual states conditioned on those CoDs. The objective is
\begin{equation}
\begin{gathered}
\mathcal{L}_{\mathrm{SFT}}
=
\mathcal{L}_{\mathrm{CoD}}
+
\lambda_{\mathrm{future}}\mathcal{L}_{\mathrm{future}},\\
\text{where}\quad
\mathcal{L}_{\mathrm{CoD}}
=
\operatorname{CE}\!\left(p_\theta(\mathbf{c}_t\mid l,\mathbf{h}_t,o_t),\mathbf{c}_t^\star\right),
\quad
\mathcal{L}_{\mathrm{future}}
=
\operatorname{dist}\!\left(\mathbf{z}_{t+\Delta},\mathbf{z}_{t+\Delta}^\star\right).
\end{gathered}
\label{eq:sft_loss}
\end{equation}
Here, $\operatorname{dist}(\cdot,\cdot)$ is a feature distance metric and $\lambda_{\mathrm{future}}$ weights the future prediction loss. For continuous control, the action expert uses the VLM representations shaped by these objectives, rather than decoding or directly consuming CoD tokens~\citep{intelligence2025pi0,chen25ecotlite,sun2026revisiting}. Unlike CoD targets, future prediction targets are available from \emph{any} temporally ordered human or robot video. In Sec.~\ref{sec:world_rl}, we use observed futures to evaluate sampled CoDs and extend learning to transitions without motion annotations.

\subsection{Scaling Chain-of-Dynamics Reasoning via Reinforcing World Imagination}
\label{sec:world_rl}

Once trained with joint supervision, the model generates CoDs and predicts future visual states conditioned on them. However, prediction under teacher-forced CoDs alone does not train the VLM to plan the intermediate dynamics needed to reach a desired future state. Therefore, we reward sampled CoDs according to the agreement between their imagined and observed futures, encouraging the VLM to reason about the world changes implied by its CoDs, as shown in Fig.~\ref{fig:architecture}(b). Since this reward requires only observed transitions, it enables the VLM to explore alternative CoDs and learn from human and robot videos without requiring action labels or executing candidate motions.

\textbf{Reward Shaping from World Imagination.}
For each candidate CoD, ThinkWorld predicts its future visual representation and compares it with the observed future in latent space. We compute the reward using a \emph{motion-enhanced feature similarity} score $\operatorname{sim}(\cdot,\cdot)$, which places greater weight on DINO feature agreement in patches with larger current-to-future changes:
\begin{equation}
    r_\mathrm{world} = \operatorname{sim}(\mathbf{z}_{t+\Delta}, \mathbf{z}_{t+\Delta}^\star).
\end{equation}
The score retains a uniform contribution from the full scene to preserve global consistency while preventing redundant background regions from dominating the reward~\citep{shah2026mv2mae}. Motion weighting uses only observed frames, so scoring requires no action annotations.

\textbf{Optimizing CoD Reasoning with World-Imagination RL.}
We aim to learn CoDs that capture the intermediate dynamics leading from the current scene to a desired future state. To train this reasoning process, ThinkWorld samples alternative CoDs from the same visual context and rewards those whose imagined futures best match the observed future. These rewards let ThinkWorld explore alternative CoDs that could lead from the current scene to the observed future, allowing passive human and robot videos to guide action reasoning without action labels or online execution of candidate motions.

We update the VLM with GRPO~\citep{shao2024deepseekmath} using normalized world-imagination rewards~\citep{liu2026gdpo}, while supervised replay preserves the motion grounding learned during SFT:
\begin{equation}
\begin{aligned}
\mathcal{L}_{\mathrm{RL}}
&=
\lambda_{\mathrm{GRPO}}
\mathcal{L}_{\mathrm{GRPO}}\!\left(r_{\mathrm{world}}\right)
+
\lambda_{\mathrm{replay}}
\mathcal{L}_{\mathrm{replay}}.
\end{aligned}
\label{eq:rl_objective}
\end{equation}
By comparing imagined and observed futures, this stage encourages the VLM to encode action-relevant dynamics in the representations later used for continuous control. All other components remain frozen during RL, and neither CoD generation nor future prediction is required at deployment. Complete optimization details are provided in Appendix~\ref{app:future-rl}.

\section{Experiment}
\label{sec:experiments}

\subsection{Implementation Details}

\textbf{Training Setup.}
For the motion tokenizer $\mathcal{T}(\cdot)$, we convert both robot actions and human wrist poses into a unified delta end-effector representation. We use Qwen3-VL-2B~\citep{bai2025qwen3} as the VLM backbone with the action expert from MolmoAct2~\citep{fang2026molmoact2}. Future queries are decoded by a Transformer decoder, with ground-truth future representations extracted using DINOv3~\citep{simeoni2025dinov3}.
Our training proceeds in two stages. During \emph{SFT and RL pre-training}, we first fully fine-tune the VLM for CoD reasoning and future prediction, and then adapt the action expert while freezing the VLM following~\cite{driess2026knowledge}. We subsequently perform RL with $K=4$ sampled trajectories, updating only the VLM while keeping the action expert frozen. We use 128 and 64 NVIDIA H100 GPUs for SFT and RL, respectively.
During \emph{Embodiment-Specific Post-Training}, we further adapt the pretrained model to each target embodiment. ThinkWorld-LIBERO is trained on the standard LIBERO training set to evaluate generalization and robustness to novel environments and perturbations, while ThinkWorld-YAM is adapted on MolmoAct2-BimanualYAM for real-robot experiments. We use 32 and 64 NVIDIA H100 GPUs for the two settings, respectively.

\textbf{Training Datasets.}
Our training corpus combines human egocentric data from EgoDex~\citep{hoque2026egodex} and EgoVerse~\citep{punamiya2026egoverse} with robot demonstrations from MolmoAct2-DROID/BimanualYAM~\citep{fang2026molmoact2}, Bridge/RT-1~\citep{o2024open}, BC-Z~\citep{jang2021bc}, and RoboCOIN~\citep{wu2025robocoin}. During SFT, we also use auxiliary VQA data to preserve visual-language capabilities and reduce catastrophic forgetting. Wrist views are included when available; otherwise, only the main view is used.

\begin{figure}[t]
    \centering
    \includegraphics[width=\linewidth]{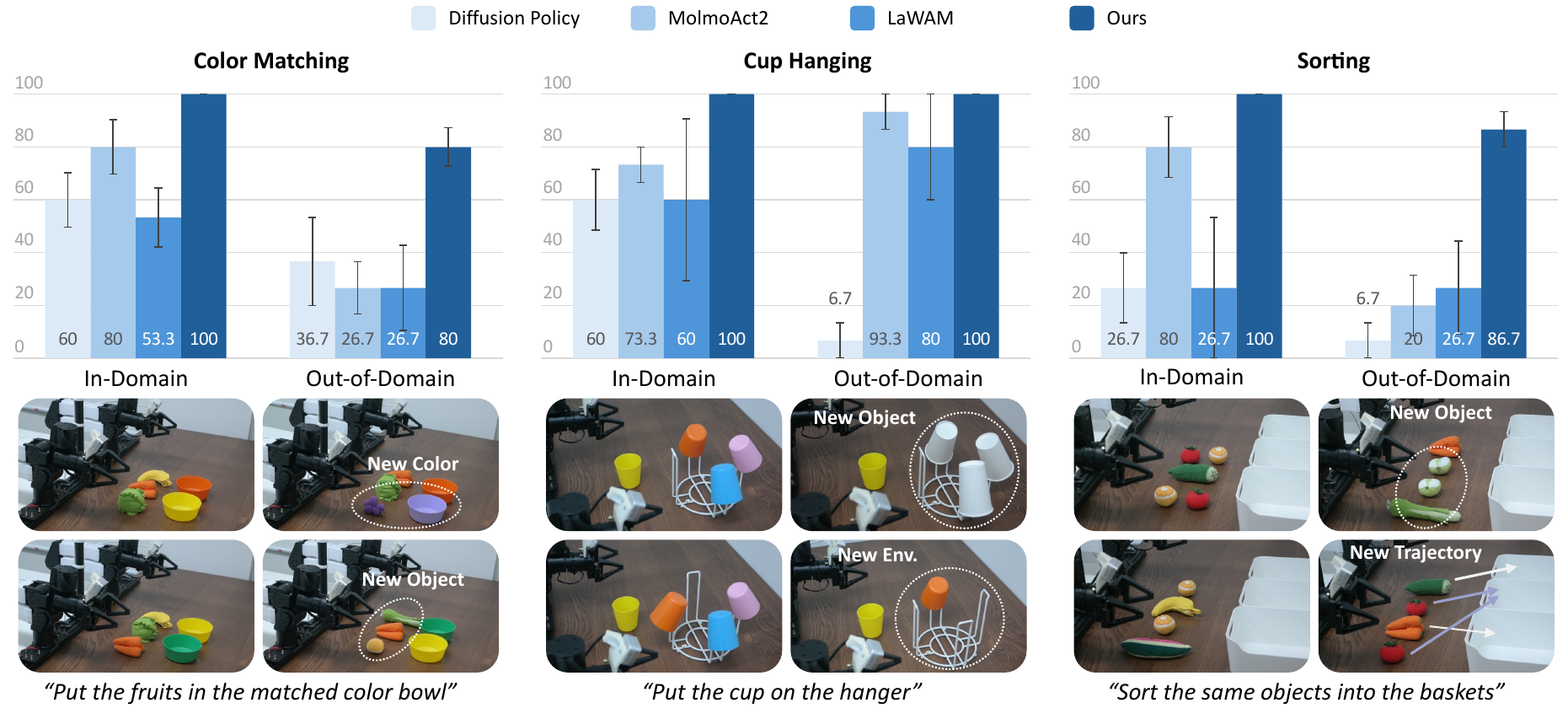}
    \caption{\textbf{Real-world out-of-distribution (OOD) generalization.} Success rates for color matching, cup hanging, and sorting under in-domain and OOD conditions. OOD settings introduce new colors, objects, environments, or sorting trajectories, as illustrated below the plots. ThinkWorld achieves the highest OOD success rates across all three tasks, despite substantial drops for the baselines.}
\label{fig:real_sr}
\end{figure}
\begin{figure}[t]
    \centering
    \includegraphics[width=\linewidth]{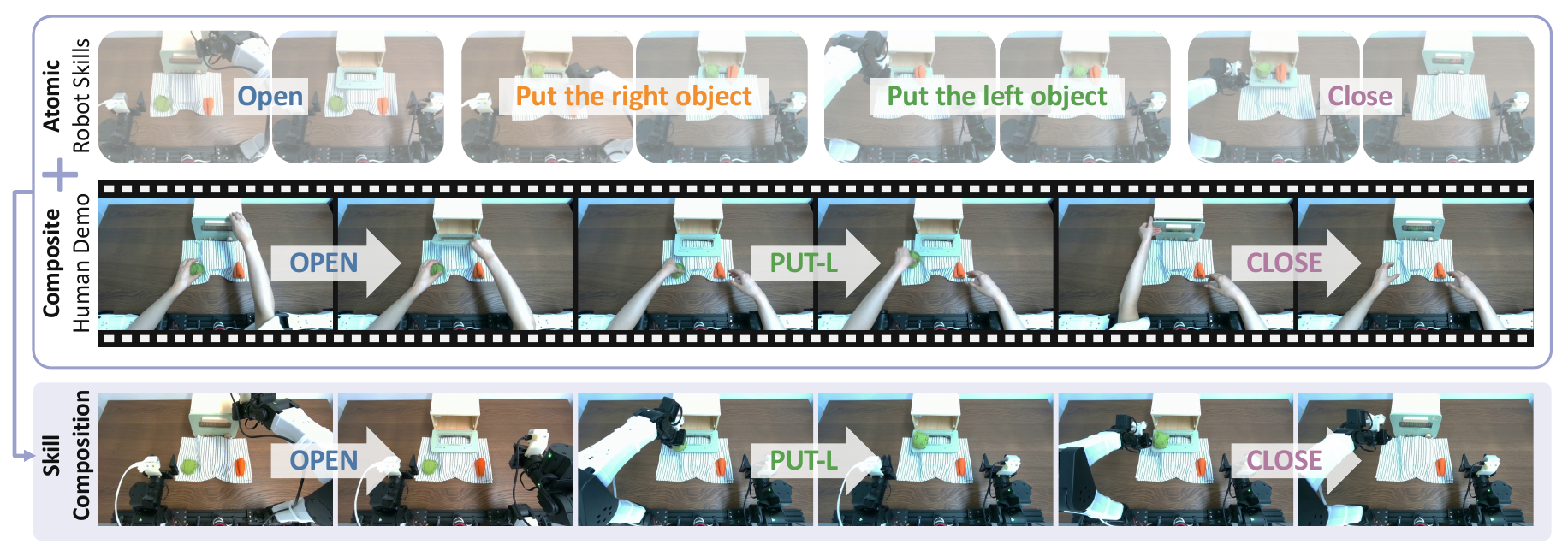}
    \vspace{-20pt}
    \caption{\textbf{Composing atomic robot skills from a human demonstration.} The \emph{top} shows atomic robot skills, while the \emph{middle} shows a human demonstration that provides the complete task composition, e.g., open $\rightarrow$ put-left $\rightarrow$ close. Guided by this human demonstration, ThinkWorld composes the corresponding robot skills and executes the complete task in a single rollout at the \emph{bottom}.}
    \label{fig:skill_composition}
\end{figure}

\textbf{Real-Robot Experimental Setup.}
We use a bimanual YAM system equipped with three cameras for real-robot evaluation: one global top-view camera and wrist-view camera for each arm. We use only RGB observations for both training and inference. The embodiment-adapted ThinkWorld-YAM model is then used as the initialization for efficient task-specific fine-tuning. 
Additional implementation and training details are provided in Appendix.


\subsection{Real-World Generalization under Distribution Shifts}

We evaluate real-world generalization under matched in-domain (ID) and out-of-distribution (OOD) settings across three manipulation tasks: placing cups on a hanger, color matching, and object sorting. OOD settings introduce changes in object color or category and, in some cases, require trajectories not observed during task-specific training. For each task, we collect demonstrations across at least six object configurations and evaluate multiple settings under both ID and OOD conditions, with five trials per setting. These tasks require the policy to identify task-relevant object attributes, such as color and category, and adapt its motion accordingly.

As shown in Fig.~\ref{fig:real_sr}, we compare ThinkWorld-YAM with Diffusion Policy~\citep{chi2023diffusion}, MolmoAct2~\citep{fang2026molmoact2}, a VLA with explicit action reasoning, and LaWAM~\citep{chen2026lawam}, which conditions action generation on predicted latent visual subgoals. While the methods perform similarly in-domain, their differences become more pronounced under OOD shifts. ThinkWorld maintains higher success rates when object appearances change or unseen trajectories are required, outperforming both MolmoAct2 and LaWAM. This suggests that grounding CoD reasoning in action-induced world changes helps the policy adapt to task-relevant changes rather than relying on appearance cues or memorized trajectories.


\subsection{Acquiring Skill Composition from Human Experience}

We further evaluate whether human demonstrations can transfer compositional knowledge to robot control. We construct an oven manipulation task from reusable atomic robot skills and provide robot demonstrations only for the individual skills \emph{1) open, 2) close, 3) put the left object in the oven}, and \emph{4) put the right object in the oven}, while a human video demonstrates how a complete compositional task can be achieved. Existing end-to-end VLA approaches typically learn executable control from robot action supervision~\citep{fang2026molmoact2}, making it difficult to compose complete robot tasks from human demonstrations when only atomic robot skills are supervised. Successful execution requires the robot to identify the relevant skills, preserve their temporal order, and transition between them without observing the complete robot trajectory during training. As shown in Fig.~\ref{fig:skill_composition}, a human demonstration of \emph{open $\rightarrow$ put-left $\rightarrow$ close} enables ThinkWorld to compose the corresponding robot skills and execute the full sequence in a single rollout. These results demonstrate that the shared CoD reasoning space enables compositional knowledge to transfer from human demonstrations to robot execution, allowing the robot to perform a complete skill sequence never demonstrated in robot data. Additional qualitative examples of skill composition in drawer manipulation are provided in the Appendix~\ref{app:real-robot-qual}.

\begin{table*}[t]
\centering
\caption{\textbf{Success rates (\%) on LIBERO and LIBERO-PRO.}
LIBERO-PRO axis values are arithmetic means over Goal, Spatial, LIBERO-10, and Object.
LIBERO-PRO baseline results are taken from the official repository unless marked with $^\dagger$, which denotes our reproduced result.
Best and second-best results within each benchmark column are shown in \textbf{bold} and \underline{underlined}, respectively.}
\label{tab:libero_and_libero_pro}
\vspace{-5pt}
\setlength{\tabcolsep}{3.5pt}
\renewcommand{\arraystretch}{1.15}

\resizebox{\textwidth}{!}{%
\begin{tabular}{
l
c
*{5}{c}
@{\hspace{10pt}}
*{6}{c}
}
\toprule
& &
\multicolumn{5}{c}{\textbf{LIBERO}}
& \multicolumn{6}{c}{\textbf{LIBERO-PRO}} \\
\cmidrule(lr){3-7}
\cmidrule(lr){8-13}

Model
& Params.
& Spatial
& Object
& Goal
& Long
& Avg.
& Object
& Position
& Semantic
& Task
& Environment
& Avg. \\
\midrule

OpenVLA~\citep{kim24openvla}
& $7.5$
& 84.7
& 88.4
& 79.2
& 53.7
& 76.5
& \underline{93.0}
& 0.0
& \textbf{97.3}
& 0.0
& \textbf{68.0}
& 51.7 \\

$\pi_0$~\citep{black2024pi_0}
& $3.2$
& 96.8
& 98.8
& 95.8
& 85.2
& 94.2
& 90.5
& 0.0
& 90.5
& 0.0
& 38.8
& 44.0 \\

$\pi_{0.5}$~\citep{intelligence2025pi0}
& $3.4$
& \textbf{98.8}
& 98.2
& \textbf{98.0}
& 92.4
& 96.9
& \textbf{96.0}
& \textbf{20.8}
& \underline{95.8}
& 0.8
& 52.8
& \underline{53.2} \\

MolmoAct~\citep{lee2025molmoact}
& $8.1$
& 87.0
& 95.4
& 87.6
& 77.2
& 86.6
& 76.0
& 1.5
& 85.8
& 1.5
& --
& 41.2 \\


VLA-JEPA~\citep{sun2026vla}
& $2.8$
& 96.2
& \textbf{99.6}
& 97.2
& \textbf{95.8}
& \underline{97.2}
& 80.4
& 3.5
& 97.2
& 20.8
& 41.0
& 48.5$^\dagger$ \\

\midrule
\rowcolor{gray!12}
\textbf{ThinkWorld (Ours)}
& \textit{$2.9$}
& \underline{98.2}
& \underline{98.8}
& \underline{97.8}
& \underline{94.2}
& \textbf{97.3}
& 73.3
& \underline{18.2}
& 93.9
& \textbf{31.9}
& \underline{60.9}
& \textbf{55.6} \\

\bottomrule
\end{tabular}%
}
\end{table*}

\subsection{Robustness to Task and Environment Shifts in Simulation}

We evaluate ThinkWorld on standard LIBERO~\citep{liu2023libero} and the more challenging LIBERO-PRO~\citep{zhou2025libero}. LIBERO evaluates manipulation across spatial, object, goal, and long-horizon task suites, while LIBERO-PRO introduces controlled shifts in object attributes, positions, instruction semantics, task goals, and environments. ThinkWorld-LIBERO is post-trained only on the standard LIBERO training set and evaluated directly on LIBERO-PRO without exposure to these perturbations. As shown in Tab.~\ref{tab:libero_and_libero_pro}, ThinkWorld achieves the best overall performance on both benchmarks, with average success rates of 97.3\% on LIBERO and 55.6\% on LIBERO-PRO. Notably, under task and environment shifts, ThinkWorld reaches 31.9\% and 60.9\%, outperforming VLA-JEPA by 11.1 and 19.9 percentage points, respectively. While VLA-JEPA also incorporates latent world modeling, these gains suggest that grounding CoD reasoning in action-induced world changes provides more transferable structure for adapting to shifts beyond the training distribution.


\begin{figure}[t]
  \centering

  \begin{minipage}[t]{0.49\linewidth}
    \centering
    \vspace{0pt}
    \includegraphics[width=\linewidth]{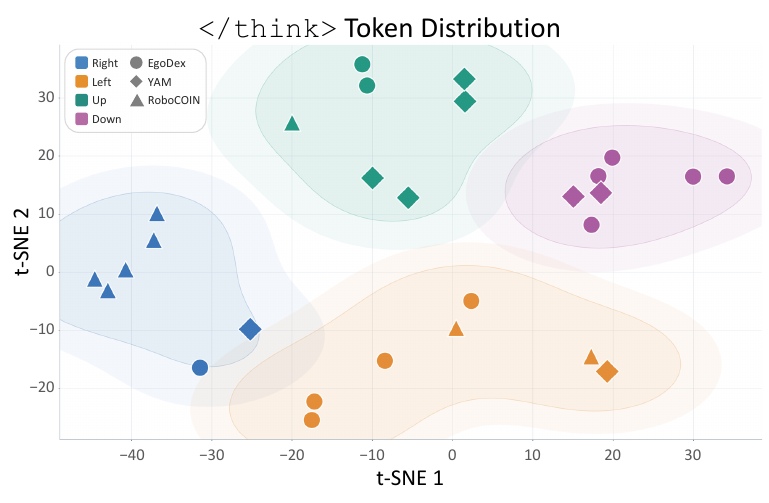}
    \vspace{-20pt}
    \caption{\textbf{t-SNE visualization of CoD across embodiments.}
    Colors denote four end-effector motion directions and marker shapes
    denote datasets. Similar motions cluster across human and robot data
    rather than by source, suggesting action structure shared across embodiments.}
    \label{fig:tsne}
  \end{minipage}\hfill
  \begin{minipage}[t]{0.49\linewidth}
    \centering
    \vspace{0pt}
    \includegraphics[width=\linewidth]{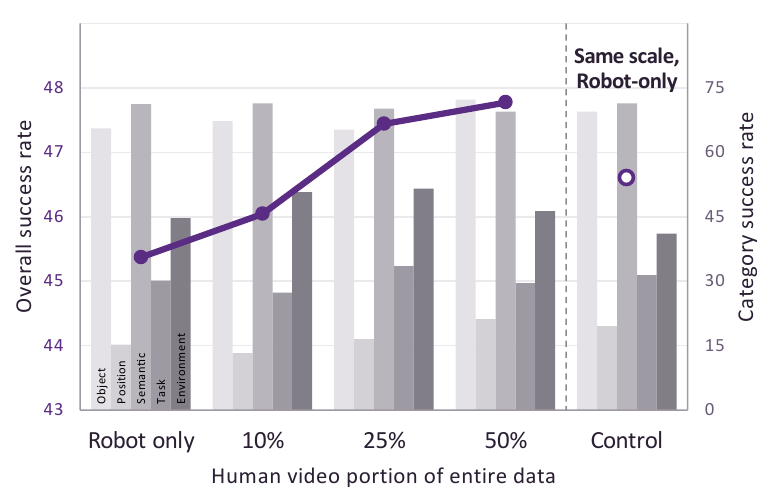}
    \vspace{-20pt}
    \caption{\textbf{Effectiveness of human data.} The LIBERO-PRO evaluation success rates (\%) on the model variants trained with different amount of human egocentric video, fixing the robot data quantity. Leveraging human data continuously enhances the performance.}
    \label{fig:scaling}
  \end{minipage}
\end{figure}
\begin{figure}[h!]
    \centering
    \includegraphics[width=\linewidth]{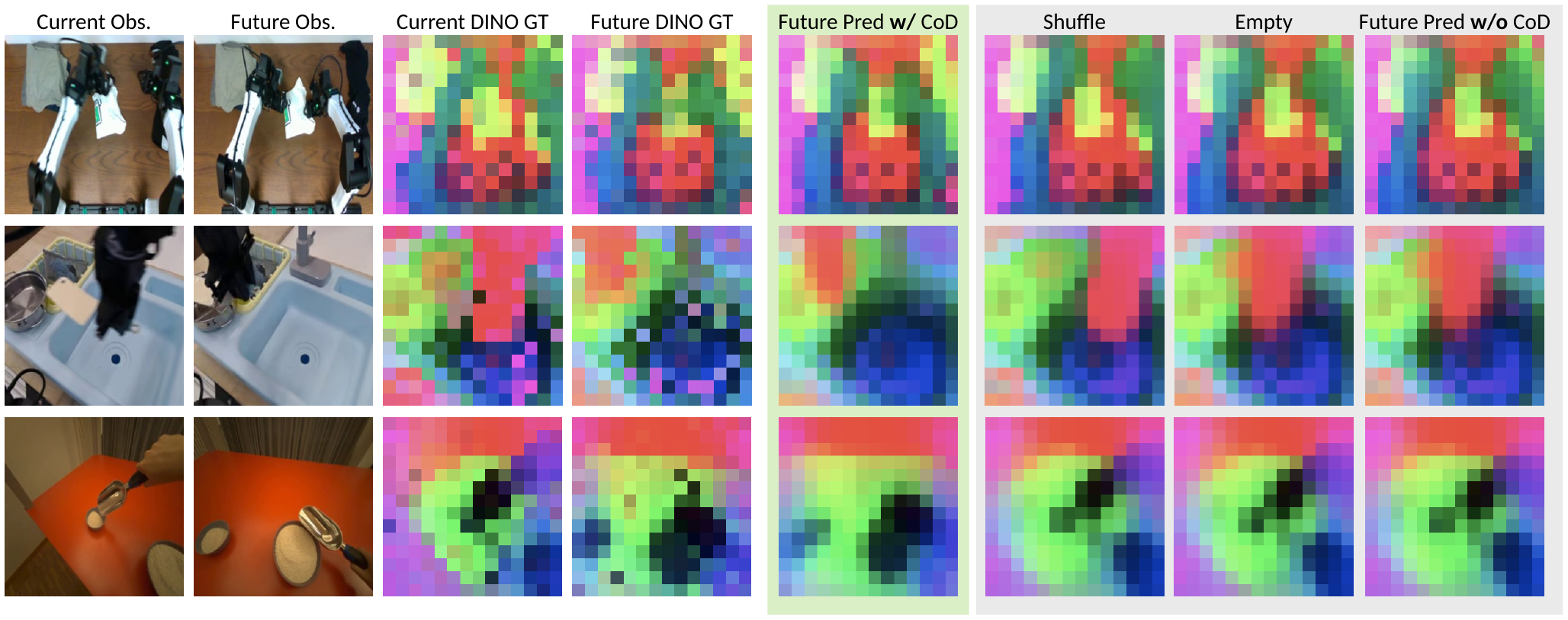}
    \vspace{-20pt}
    \caption{\textbf{Effect of Chain-of-Dynamics (CoD) on future prediction.} We compare future DINO predictions under generated, shuffled, empty, and no-CoD conditions against the current and future ground-truth features. The generated CoD captures the future-state structure, whereas the other variants largely remain similar to the current state and fail to reflect the expected future change.}
\label{fig:analysis_fp}
\end{figure}

\subsection{Analysis}

\textbf{Cross-Embodiment Structure of CoD.}
We first examine whether CoD induces a shared motion-centric structure across human and robot embodiments. For each sample, we extract the VLM hidden state of the final \texttt{</think>} token, which summarizes the generated CoD, and visualize these features with t-SNE in Fig.~\ref{fig:tsne}. We use bimanual human and robot datasets, with colors denoting four 2D end-effector motion directions and marker shapes indicating dataset sources. Despite substantial differences in embodiment and visual appearance, samples with similar motion directions form coherent clusters that contain examples from multiple datasets, while samples from the same dataset separate according to their motions. This suggests that CoD features are organized primarily around action-relevant motion rather than embodiment identity, revealing a shared action-level structure shared across human and robot behaviors.

\textbf{Effectiveness of Human Data.}
We investigate the contribution of human egocentric video through a controlled pretraining study, keeping the amount of robot data fixed while increasing the human video fraction to 10\%, 25\%, and 50\% of the total training data. We compare these variants against a robot-only baseline and an additional robot-only control that matches the training scale and number of training iterations of the 50\% human-data variant by using additional robot data. This control helps distinguish the benefits of human video from those of increased training scale alone. These experimental configurations are designed to assess the contribution of heterogeneous human video to robustness and differ from our final model configuration. After SFT, each variant undergoes embodiment-specific post-training on LIBERO and is evaluated on LIBERO-PRO to assess generalization. As shown in Fig.~\ref{fig:scaling}, overall success increases monotonically with the proportion of human video, from 45.4\% for the robot-only baseline to 47.8\% for the 50\% human-data variant. The latter also outperforms the matched robot-only control by approximately 1.2 percentage points. These results suggest that the diversity provided by human egocentric video improves generalization beyond the gains attributable to increased training scale alone.

\begin{figure}[t]
    \centering
    \includegraphics[width=\linewidth]{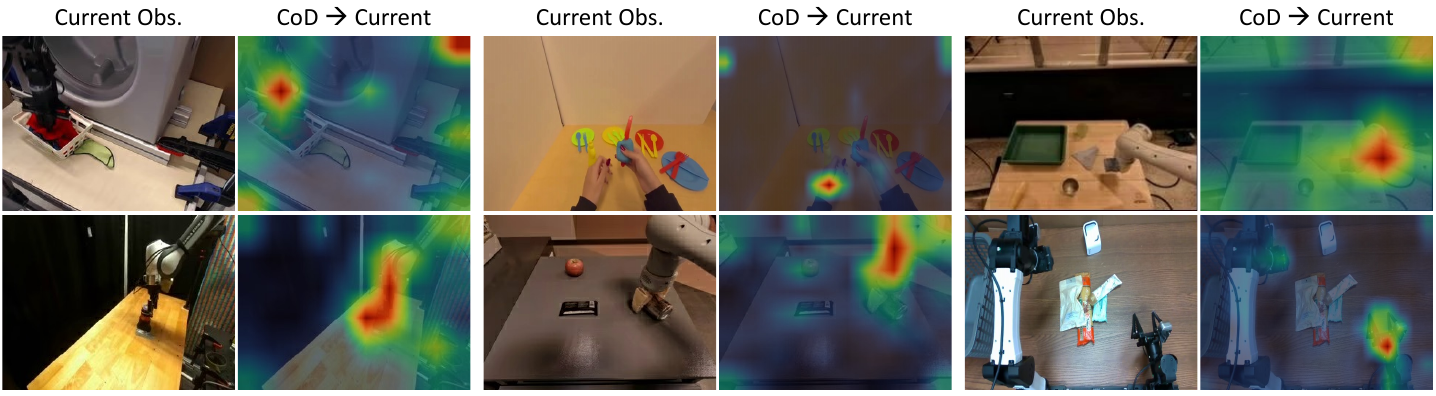}
    \caption{\textbf{Attention analysis of CoD.} We visualize attention from CoD tokens to the current observation. Across human and robot examples, CoD attends to action-relevant regions, including human hands, robot arms, and manipulated objects, highlighting its focus on motion-relevant cues.}
    \label{fig:analysis_attn}
\end{figure}
\begin{table*}[t]
\centering

\begin{minipage}[t]{0.46\textwidth}
\vspace{0pt}
\centering
\caption{\textbf{Effect of CoD on future prediction.}
Cosine similarity to ground-truth DINOv3 features one second ahead, over all patches and top-25\% motion patches. $\Delta$ is relative to CoD on; higher is better. Best in \textbf{bold}.}
\label{tab:future-prediction}

\setlength{\tabcolsep}{4pt}
\resizebox{\linewidth}{!}{%
\begin{tabular}{lcccc}
\toprule
\multirow[c]{2}{*}{Condition}
& \multicolumn{2}{c}{All patches}
& \multicolumn{2}{c}{Top-25\% motion} \\
\cmidrule(lr){2-3}\cmidrule(l){4-5}
& Cos $\uparrow$ & $\Delta$
& Cos $\uparrow$ & $\Delta$ \\
\midrule
Shuffle & 0.71 & $-0.024$ & 0.55 & $-0.071$ \\
Empty   & 0.72 & $-0.020$ & 0.54 & $-0.079$ \\
CoD off & 0.72 & $-0.018$ & 0.55 & $-0.071$ \\
\midrule
\rowcolor{gray!12}
CoD on  & \textbf{0.74} & ---
        & \textbf{0.62} & --- \\
\bottomrule
\end{tabular}%
}
\end{minipage}\hfill
\begin{minipage}[t]{0.52\textwidth}
\vspace{0pt}
\centering
\caption{\textbf{Effectiveness of training components.}
Success rates (\%)  on LIBERO-PRO when visual history frames, future prediction objective, CoD, and RL are cumulatively added to the Qwen baseline.}
\label{tab:libero_pro_component_ablation}

\setlength{\tabcolsep}{4pt}
\renewcommand{\arraystretch}{1.15}
\resizebox{\linewidth}{!}{%
\begin{tabular}{l|cccccc}
\toprule
\multirow[c]{2}{*}{Model variant}
& \multicolumn{6}{c}{\textbf{LIBERO-PRO}} \\
\cmidrule(lr){2-7}
& Object & Pos. & Sem. & Task & Env. & Avg. \\
\midrule
Qwen baseline
& 59.5 & 1.9 & 84.4 & 17.2 & 37.8 & 40.1 \\
+ History frames
& 61.3 & 4.9 & 84.6 & 22.8 & 29.8 & 40.7 \\
+ Future prediction
& 70.0 & 11.9 & 89.4 & 26.3 & 54.4 & 50.4 \\
+ CoD
& 74.5 & 16.9 & 94.4 & 29.7 & 56.4 & 54.4 \\
+ RL
& 73.3 & 18.2 & 93.9 & 31.9 & 60.9 & \textbf{55.6} \\
\bottomrule
\end{tabular}%
}
\end{minipage}

\end{table*}

\textbf{Role of CoD in Future Prediction.}
We further examine how CoD contributes to future prediction. As shown in Fig.~\ref{fig:analysis_fp}, predictions conditioned on the generated CoD better capture the action-induced transition toward the ground-truth future, whereas shuffled, empty, or no-CoD variants tend to preserve the current-state structure or predict incorrect transitions. Tab.~\ref{tab:future-prediction} quantitatively supports this observation: CoD achieves a cosine similarity of 0.74 across all patches, compared with 0.71--0.72 for the other conditions. The difference is more pronounced in the top 25\% most dynamic target patches, where CoD reaches 0.62 compared with 0.54--0.55, indicating that CoD is particularly effective at predicting action-induced changes.

We further examine what information CoD captures through attention visualization in Fig.~\ref{fig:analysis_attn}. Across both human and robot examples, CoD consistently attends to action-relevant dynamic regions, including human hands or robot arms and the manipulated objects, rather than static background regions. This pattern indicates that CoD focuses on the parts of the scene most involved in the ongoing motion and links these motion cues to the resulting changes in object state and configuration. Together with the future-prediction results, these visualizations suggest that CoD captures action-centric dynamics that ground future prediction in the state transitions induced by an action.

\textbf{Effect of Training Components.}
To assess the effectiveness of each training component, we start from a baseline consisting of the Qwen3-VL backbone with the MolmoAct2 action expert, and progressively add visual history, future prediction, CoD, and world-imagination RL. As shown in Tab.~\ref{tab:libero_pro_component_ablation}, history provides temporal context and slightly improves the average success rate from 40.1\% to 40.7\%. Future prediction brings a much larger gain to 50.4\%, indicating the benefit of explicitly modeling how the scene evolves. Adding CoD further raises performance to 54.4\%, with consistent improvements across all distribution shifts, showing the value of grounding future prediction in action-relevant dynamics. Finally, world-imagination RL improves the average to 55.6\%, with further gains under position, task, and environment shifts, demonstrating its role in refining CoD reasoning beyond supervised training. Overall, these ablations support the design of ThinkWorld, showing that each component contributes to more robust and generalizable action reasoning.

\section{Conclusion}
\label{sec:conclusion}

We present \emph{ThinkWorld}, a VLA learning framework that learns transferable Chain-of-Dynamics (CoD) through reinforcement learning with world imagination. By grounding action thoughts in the world changes they induce, ThinkWorld learns a common representation across human and robot embodiments and extends action reasoning to videos without explicit action annotations. Extensive experiments demonstrate effective human-to-robot transfer and strong generalization to unseen tasks and environments, validating the effectiveness of learning transferable action reasoning from observed world dynamics and its potential for scaling robot learning with diverse human and robot experience.

\clearpage
\appendix

\section*{Appendix}
\begin{itemize}
  \item \textbf{Section~\ref{app:implementation}. Extended Implementation Details}
  \begin{itemize}
    \item Section~\ref{app:model-configuration}. Model Configuration
    \item Section~\ref{app:io-configuration}. Input and Output Configuration
    \item Section~\ref{app:training-data}. Training Data
    \item Section~\ref{app:training-configuration}. Training Configuration
    \item Section~\ref{app:future-prediction-detail}. Details on Future Prediction Loss Design
    \item Section~\ref{app:future-rl}. Details on RL Training Objectives
  \end{itemize}
  \item \textbf{Section~\ref{app:evaluation}. Evaluation Details}
  \begin{itemize}
    \item Section~\ref{app:simulation-evaluation}. Simulation Benchmarks and Baselines
    \item Section~\ref{app:real-robot-setup}. Real-Robot Setup
    \item Section~\ref{app:real-robot-evaluation}. Real-Robot Evaluation
  \end{itemize}
  \item \textbf{Section~\ref{app:additional-results}. Additional Analyses and Results}
  \begin{itemize}
    \item Section~\ref{app:real-robot-qual}. Detailed Real-Robot Evaluation Results
    \item Section~\ref{app:more-analysis qual}. Additional Qualitative Results on PCA and Attention Analysis
    \item Section~\ref{app:simplerenv-future-qualitative}. Closed-Loop Future-Prediction Results
    \item Section~\ref{app:libero-plus}. LIBERO-Plus Evaluation Results
    \item Section~\ref{app:inference-efficiency}. Inference Efficiency
  \end{itemize}
  \item \textbf{Section~\ref{app:limitations}. Limitations and Reproducibility}
\end{itemize}

\section{Extended Implementation Details}
\label{app:implementation}

\subsection{Model Configuration}
\label{app:model-configuration}

ThinkWorld uses Qwen3-VL-2B-Instruct~\citep{bai2025qwen3} as its vision-language backbone and a frozen DINOv3 ViT-H/16+ target encoder~\citep{simeoni2025dinov3}. The DINO encoder receives the main-camera target image at $224\times224$ resolution and produces a $14\times14$ grid of 1,280-dimensional patch features. We append 64 learned future-query tokens to the VLM sequence. A two-layer MLP and a six-layer cross-attention decoder map their 2,048-dimensional hidden states to 196 predicted DINO patch features. The future target is the main-camera observation closest to exactly one second after the current observation; samples more than 50~ms from the requested time are masked from the future loss.

Continuous actions are produced by an action expert initialized from MolmoAct2~\citep{fang2026molmoact2}. The expert contains 36 non-causal Transformer blocks with hidden width 768 and operates in FP32. It exposes a maximum $30\times32$ action tensor, but a two-dimensional validity mask restricts each example to its dataset-native horizon and action channels. The expert receives layer-wise Qwen key/value features from input tokens only; generated CoD tokens and future-query tokens are not part of its direct context.

For clean normalized action $a$ and Gaussian noise $\epsilon$, we sample $z\sim\operatorname{Beta}(1,1.5)$, set $t=0.001+0.999z$, and construct $x_t=(1-t)\epsilon+ta$. The action expert regresses $a-\epsilon$ using masked mean-squared error over valid entries. We draw eight independently noised flow samples per training example and use ten Euler integration steps at inference.

\subsection{Input and Output Configuration}
\label{app:io-configuration}

\paragraph{Input prompt.}
All media precede the textual metadata in a single user turn. When temporal context is available, the order is main-view history, current main view, and the remaining current views in their dataset-defined order. History contains at most 16 main-view frames sampled at 2~Hz and ordered from oldest to newest. The current main image is not repeated in the history.

The complete VLA input/output template is shown below. Bracketed lines are optional, and the textual portion is rendered as one normalized string rather than as separate lines.

\begin{tcolorbox}[
  breakable, enhanced,
  colback=gray!5, colframe=gray!55, boxrule=0.5pt, arc=3pt,
  boxsep=3pt,
  left=14pt, right=14pt, top=12pt, bottom=12pt,
  toptitle=3pt, bottomtitle=3pt,
  middle=8pt,
  fontupper=\scriptsize\ttfamily\linespread{1.15}\selectfont,
  title=\normalfont\scriptsize\bfseries Future-prediction prompt template,
  colbacktitle=gray!20, coltitle=black,
]
\begin{verbatim}
USER MEDIA:
[main-camera history video: 0--16 frames]
<current_main_image>
[<current_wrist_image_1><current_wrist_image_2>]

USER TEXT:
The task is to {task}. The setup is {embodiment} 
[ with {native_control} control].
[The current state of the robot is {state}.]
[A {N}-frame history video is followed by the current observation image. 
| The current observation frame is provided.]
The future visual target is the main-view observation exactly one second 
after the current observation. Given these, produce a future motion plan
and then predict the future state one second in the future.

ASSISTANT:
\end{verbatim}
\end{tcolorbox}

\paragraph{Proprioceptive state encoding.}
Following MolmoAct2~\citep{fang2026molmoact2}, we encode proprioceptive state as discrete tokens. Each dataset-native state dimension is normalized using training-split statistics, clipped to $[-1,1]$, uniformly quantized into 256 bins, and serialized between dedicated state-boundary tokens. The native state width is preserved, so no common padded state vector is exposed in the prompt.

\paragraph{Media processing.}
Dataset-side resizing is disabled. Qwen retains the source aspect ratio, with a pixel-area range of 4,096--200,704 for each current image and 4,096--50,176 for each history frame. State conditioning is dropped with probability 0.10 and wrist views with probability 0.20 during supervised training. The maximum multimodal sequence length is 2,048 tokens; over-length examples are removed rather than silently truncating their supervision.

Training-time image augmentation can be summarized with the following torchvision-style transform:
\begin{tcolorbox}[
  breakable, enhanced,
  colback=gray!5, colframe=gray!55, boxrule=0.5pt, arc=3pt,
  boxsep=3pt,
  left=14pt, right=14pt, top=12pt, bottom=12pt,
  toptitle=3pt, bottomtitle=3pt,
  middle=8pt,
  fontupper=\scriptsize\ttfamily\linespread{1.15}\selectfont,
  title=\normalfont\scriptsize\bfseries Image augmentation pipeline,
  colbacktitle=gray!20, coltitle=black,
]
\begin{verbatim}
import torchvision.transforms as T

transform = T.Compose([
    T.RandomResizedCrop((height, width), scale=(0.95, 0.95)),
    T.RandomRotation(degrees=5),
    T.ColorJitter(brightness=0.2, contrast=0.2,
                  saturation=0.2, hue=0.05),
    T.RandomApply([
        T.GaussianBlur(kernel_size=5, sigma=(0.1, 1.0))
    ], p=0.2),
])
\end{verbatim}
\end{tcolorbox}
The transform parameters are sampled once and replayed across all views and time steps in an example. RL disables image augmentation so that the four sampled completions for a prompt receive identical media.

CoD tokens receive language cross-entropy supervision, whereas future-query tokens are excluded from cross-entropy and supervise only the future projector. Action SFT uses an empty \verb|<think>...</think>| wrapper and disables assistant-token cross-entropy.

\paragraph{Motion Tokenizer Training.} We train a motion tokenizer following the FAST architecture~\citep{pertsch2025fast} on one-second windows of consecutive wrist-motion deltas from robot and human trajectories. Each time step contains a 6D motion vector for one hand or a 12D vector for two hands: translation in meters and rotation as a rotation vector in radians, with gripper state excluded. We normalize translation and rotation using fixed scales of 0.3448~m and 0.9119~rad, respectively, and encode the sequences with a vocabulary of 2,048 tokens. Robot poses are expressed in the robot base frame, whereas human poses are expressed in an actor-centered frame whose $+z$ axis is gravity-aligned and whose $+x$ axis follows the actor's current yaw. Training on both domains yields a shared token space for intended wrist motion. The tokens are not executed as native robot actions; a separate continuous action expert predicts executable actions.

\subsection{Training Data}
\label{app:training-data}

VLM SFT combines robot trajectories from BimanualYAM~\citep{fang2026molmoact2}, DROID~\citep{khazatsky2024droid}, RoboCOIN~\citep{wu2025robocoin}, Bridge~\citep{walke2023bridgedata}, RT-1~\citep{o2024open}, and BC-Z~\citep{jang2021bc} with egocentric human-motion trajectories from EgoVerse~\citep{punamiya2026egoverse} and EgoDex~\cite{hoque2026egodex}. Table~\ref{tab:training-data} summarizes the frozen source scale and native action contracts without exposing dataset sampling weights. Episode and frame counts describe source data before the episode-disjoint validation holdout, and a frame denotes one trajectory timestep rather than one image per camera. For Action SFT, we exclude the 50-Hz RoboCOIN trajectories and retain only its 30-Hz trajectories.

\begin{table*}[t]
  \centering
  \caption{\textbf{Training datasets and native observation/action contracts.} K and M denote thousands and millions, respectively.}
  \label{tab:training-data}
  \small
  \resizebox{\textwidth}{!}{%
  \begin{tabular}{lrrrrll}
    \toprule
    Source & Episodes & Frames & Hours & Hz & Camera view & Action representation \\
    \midrule
    YAM & 32.2K & 76.0M & 704.1 & 30 & Top + two wrists & $30\times14$D absolute joint \\
    DROID & 74.6K & 17.8M & 328.9 & 15 & Exterior + wrist & $15\times8$D absolute joint \\
    RoboCOIN & 112.0K & 82.3M & 755.2 & 30/50 & One or three views & $30/50\times(12/14)$D absolute EEF \\
    Bridge & 53.2K & 1.9M & 105.2 & 5 & Main & $5\times7$D delta EEF \\
    RT-1 & 87.2K & 3.8M & 350.6 & 3 & Main & $3\times7$D delta EEF \\
    BC-Z & 39.4K & 5.5M & 152.0 & 10 & Main & $10\times7$D delta EEF \\
    EgoVerse & 55.6K & 124.8M & 1,155.4 & 30 & Egocentric & One-second 6D/12D motion tokens \\
    EgoDex & 314.8K & 78.3M & 724.8 & 30 & Egocentric & One-second 6D/12D motion tokens \\
    \midrule
    \textbf{Total} & \textbf{769.1K} & \textbf{390.3M} & \textbf{4,276.1} & -- & Mixed & Dataset-native \\
    \bottomrule
  \end{tabular}}
\end{table*}

Auxiliary visual-language supervision covers image and video captioning, caption question answering, spatial reasoning, pointing, counting, point tracking, robotics-oriented pointing, and robot VQA. Specifically, Pixmo-Cap~\citep{deitke2025molmo}, Molmo2-ER~\citep{fang2026molmoact2}, Molmo2 Cap-LLaVA, Molmo2 Video-Track~\citep{clark2026molmo2}, RoboPoint~\citep{robopoint}, and RoboVQA~\citep{robovqa}. These examples retain their language targets and do not contribute continuous-action or future-feature losses.

\subsection{Training Configuration}
\label{app:training-configuration}

Training consists of VLM SFT, Action SFT, and RL. All phases use fused AdamW with $(\beta_1,\beta_2)=(0.9,0.95)$, $\epsilon=10^{-8}$, weight decay 0.1, gradient-norm clipping at 1.0, and DeepSpeed ZeRO stage~1 without optimizer offload. Table~\ref{tab:training-curriculum} shows the detailed configuration by phase for direct comparison.

\begin{table*}[t]
  \centering
  \caption{\textbf{Training configuration}. Batch is the effective global batch, and allocations report representative resources used for the runs.}
  \label{tab:training-curriculum}
  \small
  \resizebox{\textwidth}{!}{%
  \begin{tabular}{lccc}
    \toprule
    Setting & VLM SFT & Action SFT & RL \\
    \midrule
    Initialization & Qwen3-VL-2B & VLM SFT & Action SFT \\
    Trainable modules & Language model + future projector & Action expert & Language model \\
    Objective & CoD/VQA CE $+3\mathcal{L}_{\mathrm{future}}$ & Masked flow matching & GRPO/GDPO + weak SFT replay \\
    Data & Robot + human motion; auxiliary VQA & Robot trajectories & Robot + human motion \\
    Iterations & 150k & 50k (5k current-view + 45k history) & 2k \\
    Learning rate & VLM $10^{-5}$; projector $10^{-4}$ & Expert $10^{-4}\!\rightarrow\!5\times10^{-5}$ & $10^{-6}$ \\
    Schedule & 5\% warmup, constant & Warmup + cosine cooldown & 50-step warmup, constant \\
    Batch & 128 & 128 & 128 \\
    Seed & 42 & 43 / 44 & 45 \\
    Allocation & $16\times8$ GPUs & $8\times8$ then $16\times8$ GPUs & $8\times8$ GPUs \\
    \bottomrule
  \end{tabular}}
\end{table*}

VLM SFT and RL execute the VLM in BF16. During Action SFT, Qwen uses BF16 autocast while the action expert is stored and optimized in FP32. For downstream LIBERO evaluation, we full-finetune the final model on demonstrations from the Object, Goal, Spatial, and LIBERO-10 suites for 50k iterations using masked flow matching, cosine learning-rate decay.

\subsection{Details on Future Prediction Loss Design}
\label{app:future-prediction-detail}

Following the main text, let $\mathbf z_{t+\Delta}$ be the predicted DINO feature map and $\mathbf z^\star_{t+\Delta}=\mathcal{E}_{\mathrm{future}}(o_{t+\Delta})$ the target from the observed future, with $\Delta=1~\mathrm{s}$. The distance used in $\mathcal L_{\mathrm{future}}=\operatorname{dist}(\mathbf z_{t+\Delta},\mathbf z^\star_{t+\Delta})$ is
\begin{equation}
\operatorname{dist}(\mathbf z,\mathbf z^\star)
=\frac{1}{|\mathcal V|}\sum_{p\in\mathcal V}
\left[
1-\cos(\mathbf z_p,\mathbf z^\star_p)
+\lambda_{\mathrm{SL1}}
\operatorname{SmoothL1}_{\beta}(\mathbf z_p,\mathbf z^\star_p)
\right],
\end{equation}
where $\mathcal V$ contains the valid image patches. Both feature maps are $\ell_2$-normalized before computing the loss. We use $\beta=0.01$ and $\lambda_{\mathrm{SL1}}=0.25\times21.792413$. During RL, the future projector is frozen and $\mathcal L_{\mathrm{future}}$ is not optimized.

\subsection{Details on RL Training Objectives}
\label{app:future-rl}

For each context, we sample $K=4$ CoDs and use the frozen future projector to obtain a predicted future representation from each candidate. The similarity function in the main-text reward $r_{\mathrm{world}}=\operatorname{sim}(\mathbf z_{t+\Delta},\mathbf z^\star_{t+\Delta})$ averages global and motion-weighted patch agreement:
\begin{equation}
\operatorname{sim}(\mathbf z,\mathbf z^\star)
=\frac{1}{2}\left[
\frac{1}{|\mathcal V|}\sum_{p\in\mathcal V}s_p
+\frac{\sum_{p\in\mathcal V}w_p s_p}
{\sum_{p\in\mathcal V}w_p}
\right],
\end{equation}
where $s_p=\cos(\mathbf z_p,\mathbf z^\star_p)$. To form $w_p$, we measure the DINO feature change between the current and observed future frames. The weight is $0.2$ for patches at or below the median change, increases linearly to $1$ at the 90th percentile, and is clipped outside that range. For nearly static targets, we use the global average for both terms.

A malformed CoD receives a reward $0.05$ below the lowest candidate similarity in its group. We standardize rewards within each four-candidate group and then across the rollout batch; each candidate's resulting advantage is shared by its tokens. GRPO uses token-level importance ratios clipped to $[0.8,1.28]$ and a KL coefficient of $0.04$. In the main-text objective $\mathcal L_{\mathrm{RL}}$, we set $\lambda_{\mathrm{GRPO}}=0.999$ and $\lambda_{\mathrm{replay}}=0.001$. The replay term is teacher-forced cross-entropy on a separate minibatch of canonical CoDs. Only the VLM is updated during RL; the action expert and future projector remain frozen.

\section{Evaluation Details}
\label{app:evaluation}

\subsection{Simulation Benchmarks and Baselines}
\label{app:simulation-evaluation}

We fine-tune our policy only on standard LIBERO~\citep{liu2023libero} and use its unmodified tasks as the in-distribution reference. We then evaluate the trained policy on LIBERO-PRO and LIBERO-Plus without further adaptation to either benchmark. We select LIBERO-PRO as the primary benchmark because its controlled changes to objects, object positions, task specifications, and environments directly test whether the policy can adapt its behavior to new manipulation settings. LIBERO-Plus additionally covers camera, lighting, background, and sensor-noise perturbations, which probe perceptual robustness beyond our main methodological focus. We report a supplementary LIBERO-Plus evaluation in Appendix~\ref{app:libero-plus}.

\subsection{Real-Robot Setup}
\label{app:real-robot-setup}

\begin{wrapfigure}{r}{0.35\columnwidth}
  \centering
  \vspace{-5pt}
  \includegraphics[width=0.9\linewidth]{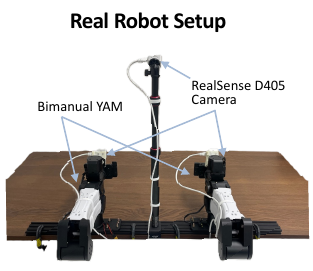}
  \vspace{-8pt}
  \caption{\textbf{BimanualYAM setup.}}
  \label{fig:suppl_robot_setup}
\end{wrapfigure}

We use the BimanualYAM robot system to collect our in-lab dataset and evaluate trained policies. As shown in Fig.~\ref{fig:suppl_robot_setup}, the system comprises two YAM arms and three RealSense D405 cameras: one providing a top view and one mounted on each wrist. Policies use RGB observations only.

Policy inference runs on a single NVIDIA RTX~3090. To account for differences in inference latency, we query each policy and execute its actions at the same 15~Hz control rate, while the remaining system components run at 30~Hz. For each predicted action chunk, we execute the first half and then replan using the latest observation.

\subsection{Real-Robot Evaluation}
\label{app:real-robot-evaluation}

For the real-robot experiments, we evaluate three tasks on the bimanual YAM platform: color matching (60 robot demonstrations), cup hanging (45), and object sorting (100). Within each task, the demonstrations span multiple trajectories and environment configurations, and we use them jointly to fine-tune a single task-specific policy. The tasks are designed to test whether the policy can infer task-level regularities and adapt its actions to the current scene rather than memorize a single trajectory. Separately, we study skill composition on the oven task using 100 robot episodes of atomic-action trajectories and 90 human-video episodes of full task sequences. We co-train a single model on the combined robot and human datasets. Example task trajectory for four tasks can be found in Fig.~\ref{fig:suppl_real_robot_qual} (a) and Fig.~\ref{fig:skillcomp_qual} (a).

We fine-tune ThinkWorld for 50,000 steps, applying LoRA to the VLM and fully updating the action DiT decoder. ThinkWorld can also leverage human videos during training; for these samples, we optimize only the VLM future-prediction loss and omit the action DiT loss. We train each baseline for 50,000 steps using its model-specific fine-tuning recipe from the official LeRobot~\citep{cadene2024lerobot} repository.

\begin{table*}[t]
  \centering
  \caption{\textbf{Full real-robot success-rate (SR) comparison.} Each task is evaluated under
  in-distribution (ID) and out-of-distribution (OOD) settings; cells report
  successes/trials with SR (\%) in parentheses. Task~1 comprises six ID and six OOD
  conditions ($5$ trials each); Tasks~2 and~3 comprise three of each. Macro SR is the
  unweighted mean of the six per-setting rates; Micro SR pools all $120$ trials.}
  \label{tab:real-robot-sr}
  \small
  \resizebox{\textwidth}{!}{%
  \begin{tabular}{lcccccccc}
    \toprule
    & \multicolumn{2}{c}{Color Matching} & \multicolumn{2}{c}{Cup Hanging} & \multicolumn{2}{c}{Sorting} & \multicolumn{2}{c}{Overall} \\
    \cmidrule(lr){2-3} \cmidrule(lr){4-5} \cmidrule(lr){6-7} \cmidrule(lr){8-9}
    Model & ID & OOD & ID & OOD & ID & OOD & Macro SR & Micro SR \\
    \midrule
    Diffusion Policy  & 18/30 (60.0)          & 11/30 (36.7)         & 9/15 (60.0)           & 1/15 (6.7)            & 4/15 (26.7)           & 1/15 (6.7)           & 32.8          & 36.7 \\
    MolmoAct2         & 24/30 (80.0)          & 8/30 (26.7)          & 11/15 (73.3)          & 14/15 (93.3)          & 12/15 (80.0)          & 3/15 (20.0)          & 62.2          & 60.0 \\
    LaWAM             & 16/30 (53.3)          & 8/30 (26.7)          & 9/15 (60.0)           & 12/15 (80.0)          & 4/15 (26.7)           & 4/15 (26.7)          & 45.6          & 44.2 \\
    \textbf{ThinkWorld (ours)} & \textbf{30/30 (100.0)} & \textbf{24/30 (80.0)} & \textbf{15/15 (100.0)} & \textbf{15/15 (100.0)} & \textbf{15/15 (100.0)} & \textbf{13/15 (86.7)} & \textbf{94.4} & \textbf{93.3} \\
    \bottomrule
  \end{tabular}}
\end{table*}

\clearpage

\section{Additional Analyses and Results}
\label{app:additional-results}



\subsection{Detailed Real-Robot Evaluation Results}
\label{app:real-robot-qual}

\textbf{In-domain and out-of-domain task evaluation.} 
Tab.~\ref{tab:real-robot-sr} reports detailed rollout success rates under in-distribution (ID) and out-of-distribution (OOD) conditions for all tasks. ThinkWorld achieves the highest success rate in all six task--condition pairs, succeeding in all 60 ID trials and 52 of 60 OOD trials. Its macro and micro success rates are 94.4\% and 93.3\%, respectively, compared with 62.2\% and 60.0\% for the strongest baseline, MolmoAct2. Example in-domain and out-of-domain rollout trajectories for all compared methods are shown in Figs.~\ref{fig:suppl_real_robot_qual} (b) and (c) respectively.

\textbf{Skill composition.} 
Fig.~\ref{fig:skillcomp_qual} presents qualitative results for real-robot skill composition in oven and drawer manipulation tasks. Each task involves opening the oven or a specified drawer, optionally placing the left or right object inside, and closing it. Robot demonstrations cover only the individual atomic skills (OPEN, PUT-L, PUT-R, and CLOSE; panel~(a)), while human demonstrations specify complete skill sequences (panel~(b)). The oven tasks include \emph{open $\rightarrow$ close} and \emph{open $\rightarrow$ put-left $\rightarrow$ close}. The drawer tasks include \emph{open $\rightarrow$ put-left $\rightarrow$ close} and \emph{open $\rightarrow$ put-right $\rightarrow$ close} for the upper drawer, and \emph{open $\rightarrow$ put-left $\rightarrow$ close} for the lower drawer. The successful rollouts in panel~(c) show ThinkWorld executing these compositions under the corresponding natural-language instructions, manipulating the specified objects and compartments while preserving the required skill order. These qualitative results illustrate the transfer of compositional knowledge from human demonstrations to robot execution without complete robot task trajectories during training.

\subsection{Additional Qualitative Results on PCA and Attention Analysis}
\label{app:more-analysis qual}

We provide additional qualitative visualizations that extend the analysis in the main text.
Future-representation PCA and attention visualizations leverage Qwen layers 24--27. PCA and attention are descriptive analyses and are not used as causal evidence.

Fig.~\ref{fig:fp_qual} expands the PCA visualization of future-prediction representations across five CoD conditions: the original CoD, shuffled CoD, empty CoD, a donor CoD from another episode, and no CoD. This comparison illustrates how the predicted future representations vary with CoD conditioning.

\begin{figure*}[p]                 
  \centering
  \includegraphics[width=\textwidth]{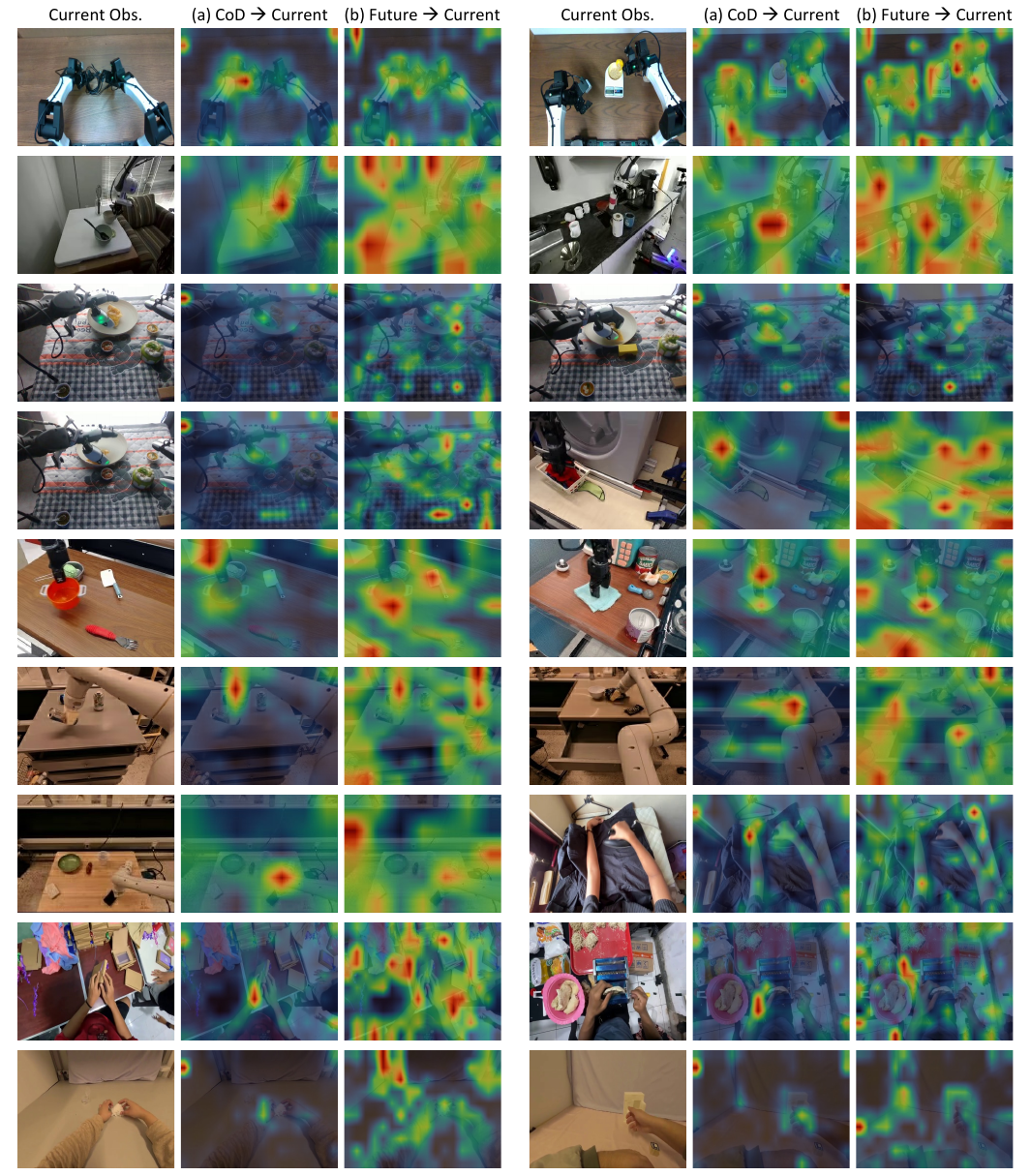}
  \caption{\textbf{Additional attention visualizations.} Attention maps from CoD and future-prediction tokens to the input observation. CoD attention concentrates on dynamic, task-relevant regions, while future-prediction attention is more broadly distributed across the scene.}
  \label{fig:suppl_attn_qual}
\end{figure*}

Fig.~\ref{fig:suppl_attn_qual} extends the CoD-to-input-observation attention maps shown in the main text and additionally shows attention from future-prediction tokens to the input observation. CoD attention tends to concentrate on dynamic, task-relevant regions, whereas future-prediction attention is distributed more broadly across the scene. This pattern suggests that CoD emphasizes localized changes relevant to action, while future prediction incorporates wider scene context.

\subsection{Closed-Loop Future-Prediction Results}
\label{app:simplerenv-future-qualitative}


\begin{wraptable}{r}{0.36\linewidth}
  \centering
  \vspace{-10pt}
  \caption{\textbf{Out-of-the-box performance on SimplerEnv.}}
  \label{tab:simplerenv-zeroshot}
  \footnotesize
  \setlength{\tabcolsep}{2pt}
  \begin{tabularx}{\linewidth}{@{}lXr@{}}
    \toprule
    Group & Task & SR (\%) \\
    \midrule
    Google-VM & & 49.31 \\
    \midrule
    \multirow{5}{*}{Google-VA}
      & Pick Coke can & 56.00 \\
      & Drawer & 33.33 \\
      & Move-near & 55.83 \\
      & Put-in-drawer & 2.12 \\
    \cmidrule(l){2-3}
      & \textbf{Micro avg.} & \textbf{46.54} \\
    \midrule
    WidowX VM & & 41.67 \\
    \midrule
    \rowcolor{gray!12}
    \textbf{Overall} & \textbf{micro} & \textbf{46.51} \\
    \bottomrule
  \end{tabularx}
\end{wraptable}

Tab.~\ref{tab:simplerenv-zeroshot} reports SimplerEnv~\citep{li2024evaluating} success rates without additional embodiment-specific fine-tuning for this benchmark. The model nevertheless encountered Bridge and RT-1 data during SFT and RL, providing prior experience with related robot settings. It achieves 49.31\% on Google visual matching, 46.54\% on the Google-VA variants (micro average), and 41.67\% on WidowX visual matching, with an overall micro success rate of 46.51\%. These results show that the policy can complete a substantial fraction of SimplerEnv rollouts without benchmark-specific adaptation.

\begin{figure}[p]
    \centering
    \includegraphics[width=\linewidth]{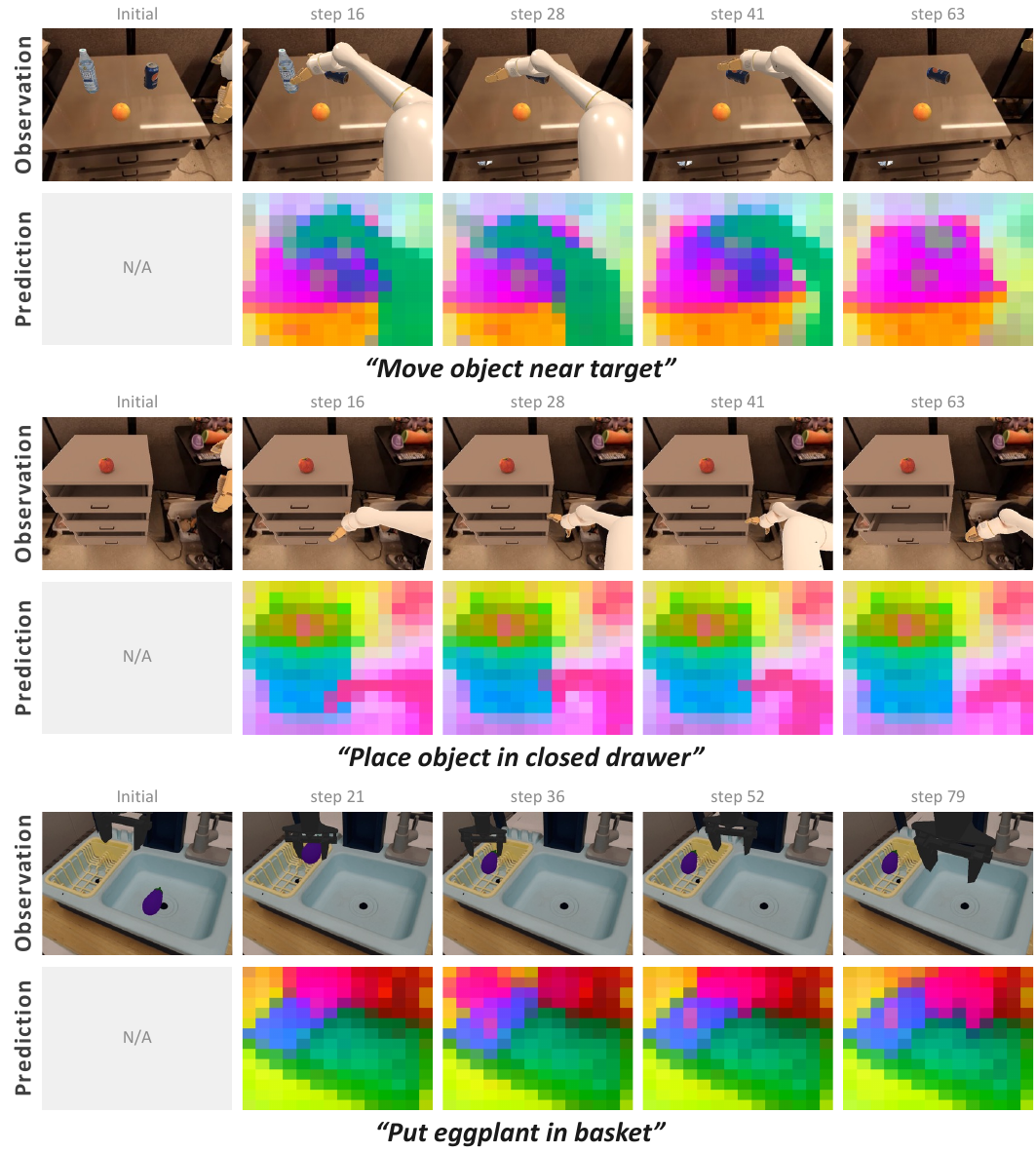}
    \caption{\textbf{One-second-ahead future predictions during closed-loop SimplerEnv rollouts.} For each task, the upper row shows observations actually reached by the policy, while the lower row shows PCA-projected future representations predicted at the preceding replanning horizon. Each column pairs a prediction with the observation at its target time, so their scene structure should correspond. The close qualitative alignment in layout and task progress shows that the predictions track the states encountered during closed-loop execution. The initial frame has no preceding prediction.}    
    \label{fig:simplerenv_fp_viz}
\end{figure}

Fig.~\ref{fig:simplerenv_fp_viz} visualizes future predictions during closed-loop SimplerEnv rollouts. In each column, the one-second-ahead prediction is paired with the observation subsequently reached at the predicted time. The predictions are PCA-projected feature maps and their spatial structure and progression closely track the corresponding observations across the three examples. This qualitative alignment suggests that the future representation remains informative as the policy encounters its own observations and replans under distribution shift.

\begin{table*}[t]
\centering
\caption{\textbf{Success rates (\%) on LIBERO-Plus across seven perturbation dimensions.} Only methods with published LIBERO-Plus results are listed. Higher is better.}
\label{tab:libero_plus}

\footnotesize
\setlength{\tabcolsep}{3pt}
\renewcommand{\arraystretch}{1.15}
\begin{tabular}{l*{8}{c}}
\toprule
Method & Camera & Robot & Language & Light & Background & Noise & Layout & Avg. \\
\midrule
\shortstack[l]{OpenVLA-OFT\\\citep{kim24openvla}}
& 56.4 & 31.9 & 79.5 & 88.7 & 93.3 & 75.8 & 74.2 & 69.6 \\

\shortstack[l]{$\pi_0$-FAST\\\citep{black2024pi_0}}
& 65.1 & 21.6 & 61.0 & 73.2 & 73.2 & 74.4 & 68.8 & 61.6 \\

\shortstack[l]{DreamVLA\\\citep{zhang2026dreamvla}}
& 65.0 & 40.8 & 63.5 & 85.7 & 82.6 & 84.9 & 74.0 & 69.9 \\
\midrule
\rowcolor{gray!12}
\textbf{ThinkWorld (Ours)}
& 46.9 & 70.5 & 73.5 & 90.7 & 90.0 & 62.7 & 79.6 & \textbf{71.7} \\
\bottomrule
\end{tabular}
\end{table*}

\subsection{LIBERO-Plus Evaluation Results}
\label{app:libero-plus}

Tab.~\ref{tab:libero_plus} reports zero-shot LIBERO-Plus~\citep{fei2026liberoplus} results using the policy fine-tuned only on standard LIBERO. ThinkWorld achieves the highest reported Avg. among the four methods at 71.7\%, exceeding DreamVLA by 1.8 percentage points. It leads under robot initial-state, lighting, and object-layout perturbations. In particular, ThinkWorld reaches 70.5\% on robot initial-state changes, compared with 40.8\% for the next-best method, and 79.6\% on object-layout changes, compared with 74.2\%. These results are consistent with our goal of adapting actions to changed physical configurations and provide supplementary evidence of generalization beyond standard LIBERO.

\subsection{Inference Efficiency}
\label{app:inference-efficiency}

\begin{wraptable}{r}{0.52\textwidth}
  \vspace{-\intextsep}
  \centering
\caption{\textbf{Inference efficiency.} Per-call latency percentiles, total wall
time, and peak GPU memory, measured at steady state over 50 consecutive
eager-mode calls on a single NVIDIA H100 GPU. Lower is better for all metrics.}
  \label{tab:inference-efficiency}
  \footnotesize
  \setlength{\tabcolsep}{3.5pt}
  \renewcommand{\arraystretch}{1.15}
  \begin{tabular}{
    lcccc
  }
    \toprule
    Policy & {p50} & {p95} & {Wall} & {Mem.} \\
           & {(ms)} & {(ms)} & {(s)} & {(GiB)} \\
    \midrule
    MolmoAct2       & 335.07 &  467.96 & 18.44 & 12.11 \\
    MolmoAct2-Think & 879.37 & 1014.47 & 46.76 & 12.12 \\
    \midrule
    \rowcolor{gray!12}
    \textbf{ThinkWorld}
      & \bfseries 324.84
      & \bfseries 333.09
      & \bfseries  16.31
      & \bfseries   7.07 \\
    \bottomrule
  \end{tabular}
  \vspace{-\intextsep}
\end{wraptable}

Table~\ref{tab:inference-efficiency} reports steady-state batch-1 policy latency with CUDA graph capture disabled, using eager inference without static-shape graph setup. All models are measured on the same NVIDIA H100 80GB GPU, with five warm-up calls followed by 50 CUDA-synchronized timed calls. Each model uses its native deployed input contract, so the table reflects practical policy latency rather than an identical-input architecture ablation.

ThinkWorld uses the reasoning process only during training, where it encourages the VLM to learn more action-centric representations. The deployed policy does not execute this process at inference, so each policy call avoids its additional runtime computation. This separation offers an explanation for how training-time reasoning can support strong task performance while preserving low inference latency.

\section{Limitations and Reproducibility}
\label{app:limitations}

Training results are obtained from a single training seed; accordingly, the reported episode-level confidence intervals characterize evaluation variability rather than variation across independently trained checkpoints. For real-robot experiments, we report the number of physical trials and task-level results to make the evaluation scope explicit. Inference latency should be interpreted under the stated deployment setup, as it varies with hardware and input configuration. To support reproducibility, we provide trial denominators, model initialization details, training configurations, and hardware specifications alongside the aggregate results.


\newcommand{\rolloutcaption}{\textbf{Real-robot experiment task rollout results.}
(a) Sample teleoperation demonstrations used for training.
(b) In-domain rollouts. (c) Out-of-domain rollouts.
Each panel shows qualitative rollout quality for every method,
with frames sampled uniformly over an episode and time advancing
left to right; the corresponding success rates are reported in Tab.~\ref{tab:real-robot-sr}.}

\begin{figure}[p]
  \centering
  \includegraphics[width=\textwidth,height=0.88\textheight,keepaspectratio]{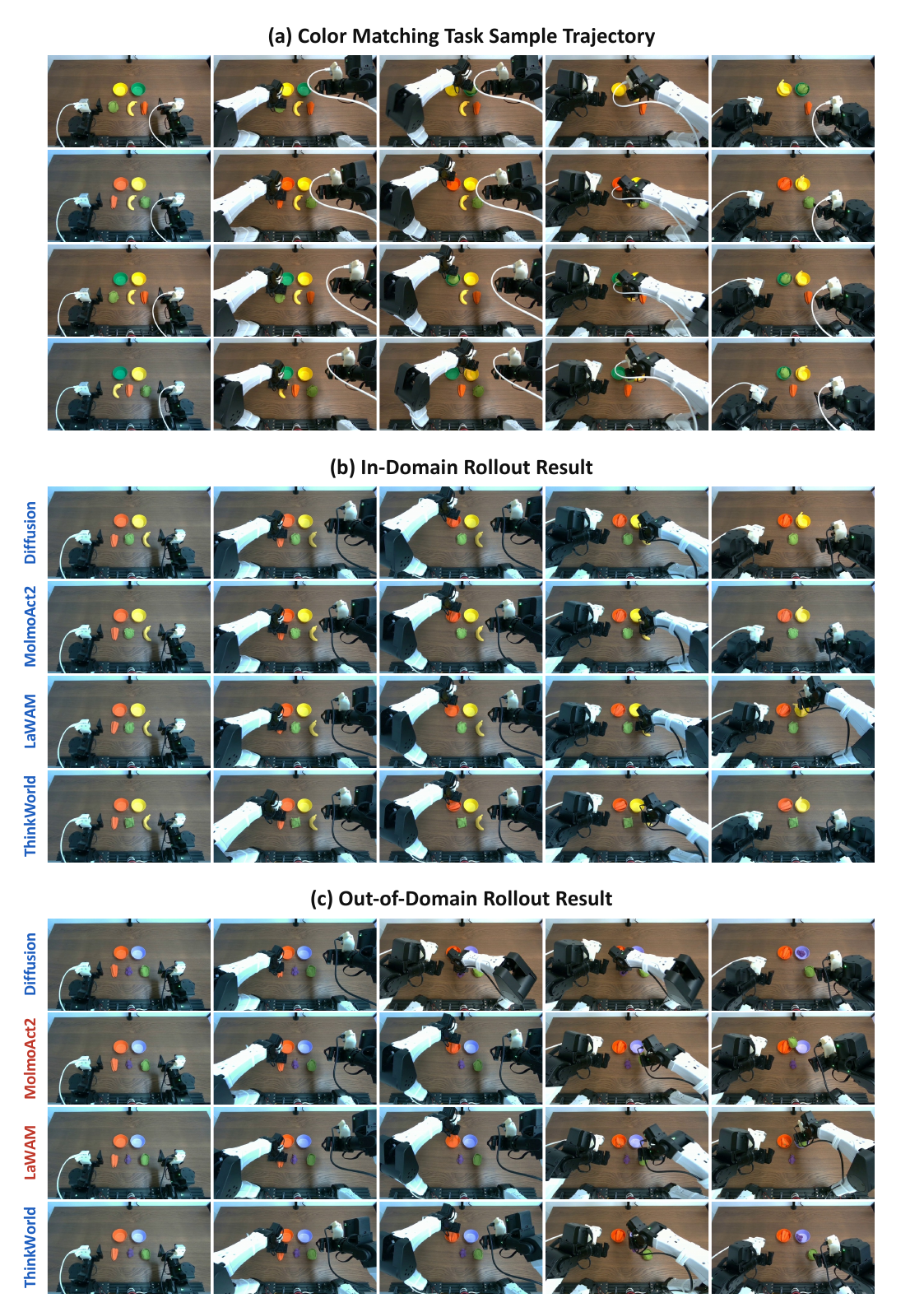}
  \caption{\rolloutcaption}
  \label{fig:suppl_real_robot_qual}
\end{figure}

\begin{figure}[p]\ContinuedFloat
  \captionsetup{list=no}
  \centering
  \includegraphics[width=\textwidth,height=0.88\textheight,keepaspectratio]{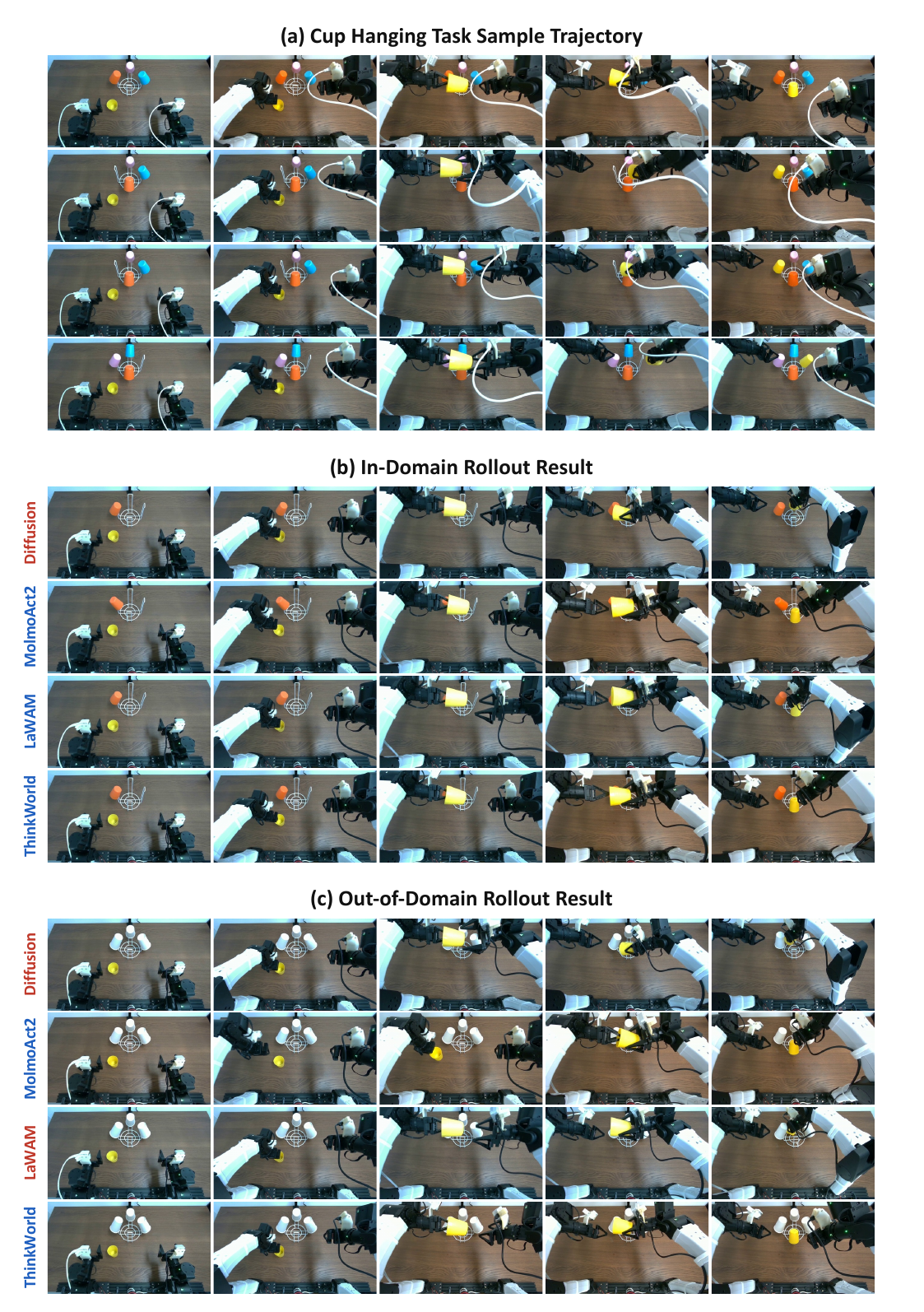}
  \caption{\rolloutcaption}
\end{figure}

\begin{figure}[p]\ContinuedFloat
  \captionsetup{list=no}
  \centering
  \includegraphics[width=\textwidth,height=0.88\textheight,keepaspectratio]{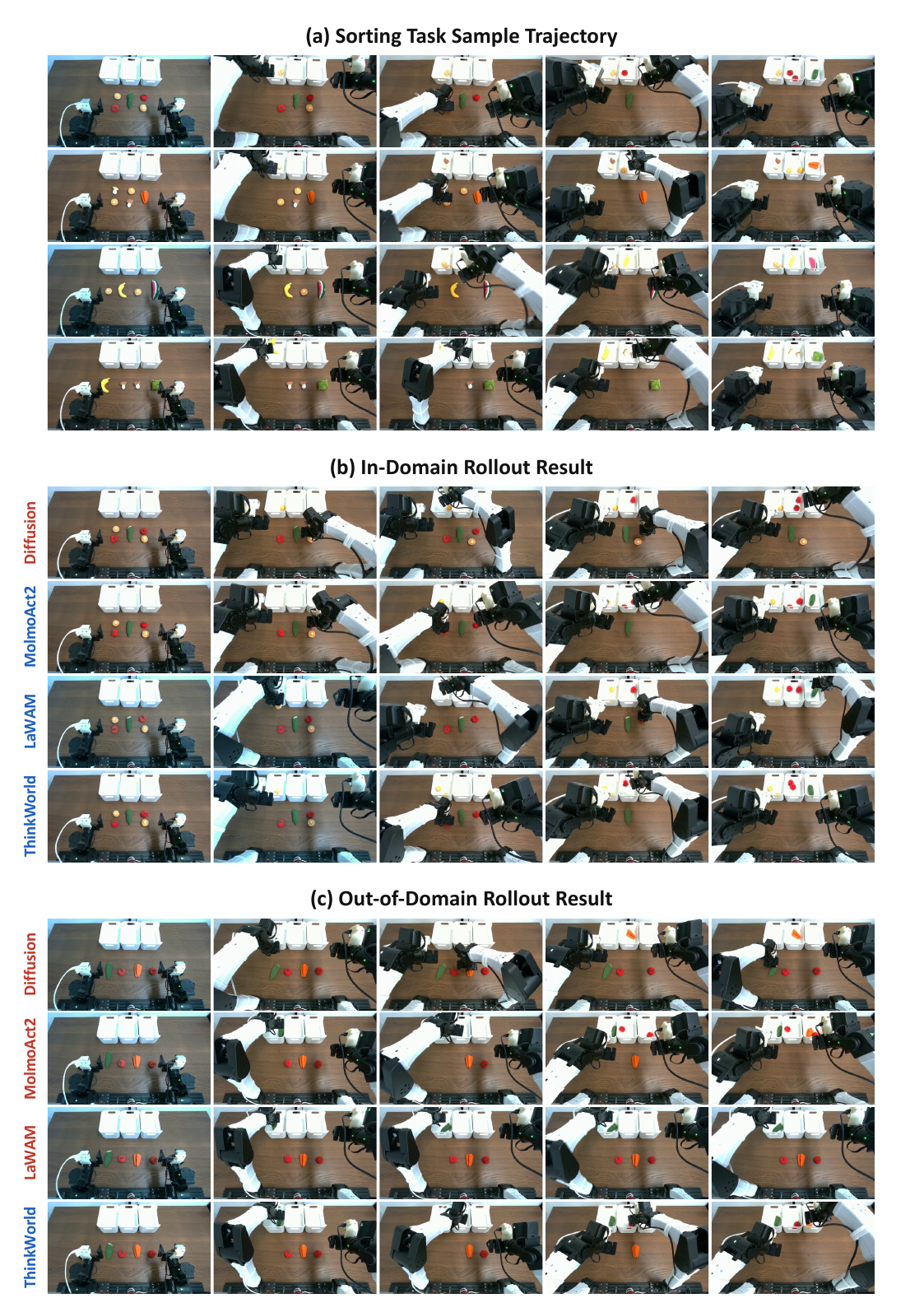}
  \caption{\rolloutcaption}
\end{figure}
\begin{figure}[p]
    \centering
    \includegraphics[width=0.7\linewidth]{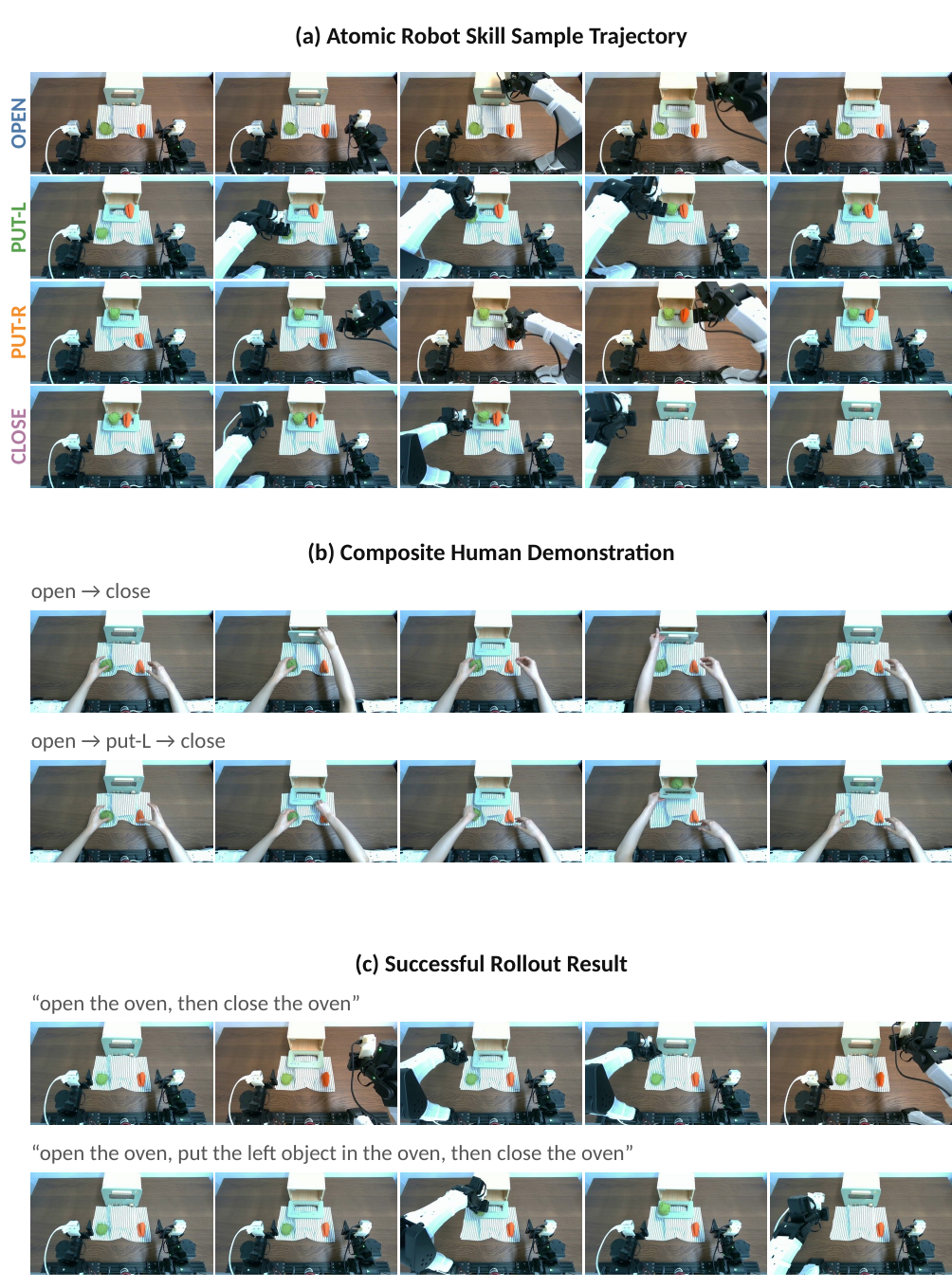}
    \caption{\textbf{Real-Robot skill composition sample trajectory and rollout results.}
    (a) Sample atomic robot trajectories for opening the oven (OPEN), placing the left or right object inside (PUT-L, PUT-R), and closing the oven (CLOSE).
    (b) Human demonstrations that combine these skills into open $\rightarrow$ close and open $\rightarrow$ put-L $\rightarrow$ close sequences.
    (c) Successful robot rollouts for the corresponding natural-language instructions.
    Frames in each row progress from left to right.}
    \label{fig:skillcomp_qual}
\end{figure}

\begin{figure}[p]
    \ContinuedFloat
    \centering
    \includegraphics[width=0.6\linewidth]{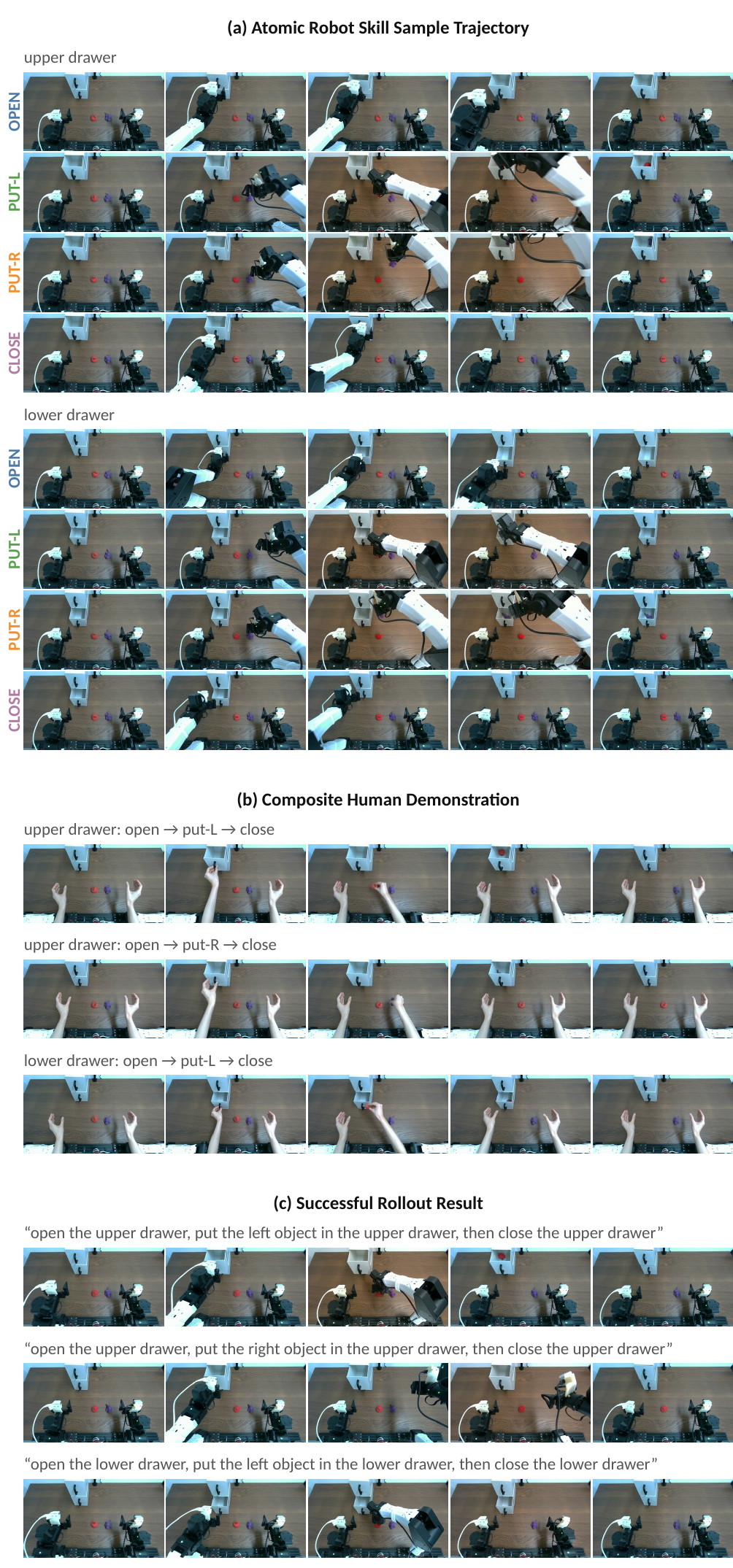}
    \caption[]{\textbf{Real-Robot skill composition sample trajectory and rollout results.}
    (a) Sample atomic robot trajectories for opening the upper or lower drawer (OPEN), placing the left or right object inside (PUT-L, PUT-R), and closing the drawer (CLOSE).
    (b) Human demonstrations that combine these skills into open $\rightarrow$ put-L $\rightarrow$ close and open $\rightarrow$ put-R $\rightarrow$ close sequences for the upper drawer, and an open $\rightarrow$ put-L $\rightarrow$ close sequence for the lower drawer.
    (c) Successful robot rollouts for the corresponding natural-language instructions.
    Frames in each row progress from left to right.}
\end{figure}
\begin{figure*}[p]
  \centering
  \includegraphics[width=\textwidth]{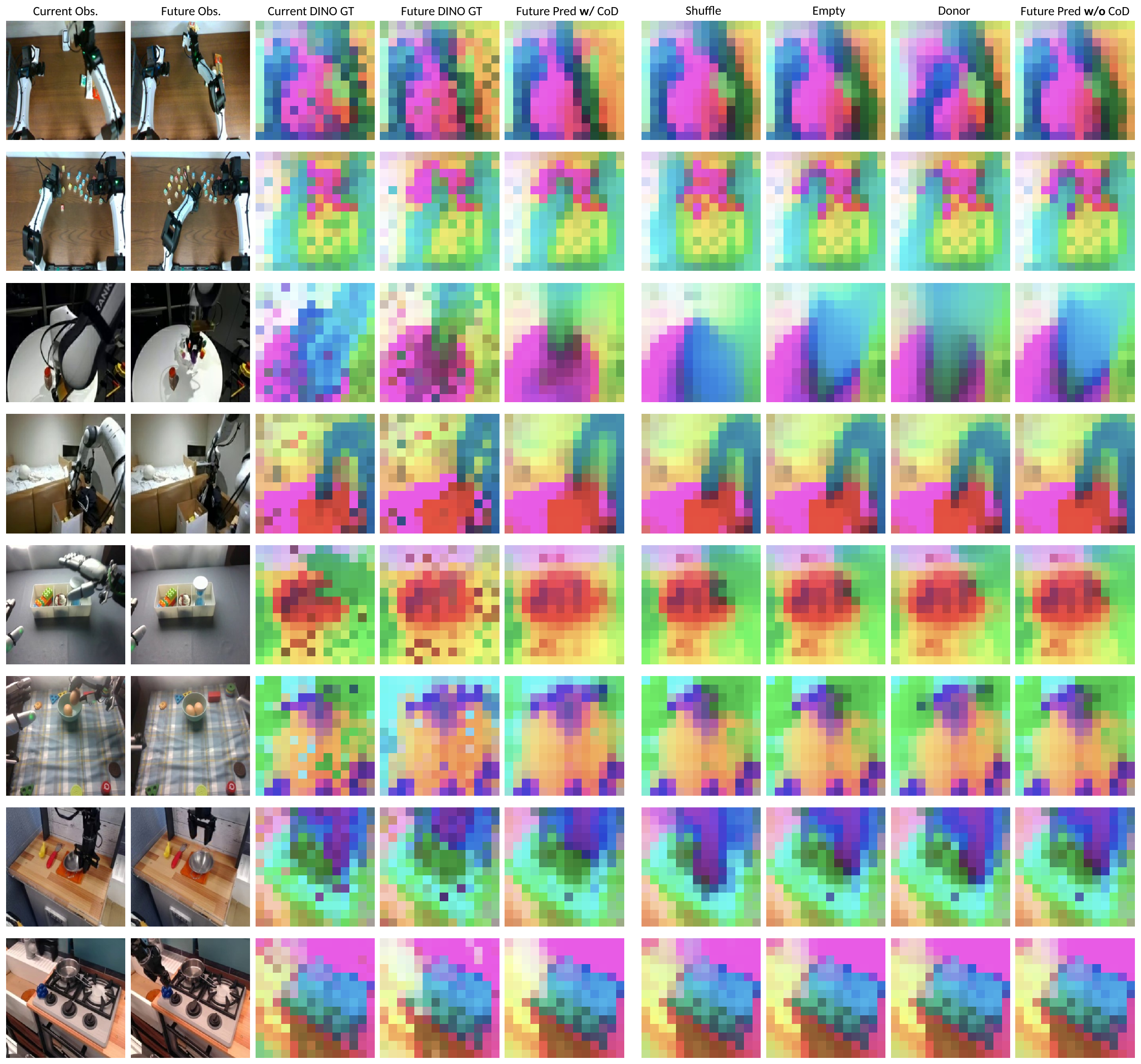}
  \caption{%
    \textbf{Qualitative future-prediction results.}
    Each row is one episode.
    The first two columns show the current and ground-truth future RGB
    observations, and the next two show PCA projections of the
    ground-truth DINO features for those frames.
    The remaining five columns show the predicted future DINO features
    under different CoD conditions: with CoD, shuffled CoD, empty CoD,
    a CoD taken from a different (donor) episode, and without CoD.
    All PCA panels share one color mapping.%
  }
  \label{fig:fp_qual}
\end{figure*}

\begin{figure*}[p]\ContinuedFloat
  \captionsetup{list=no}
  \centering
  \includegraphics[width=\textwidth]{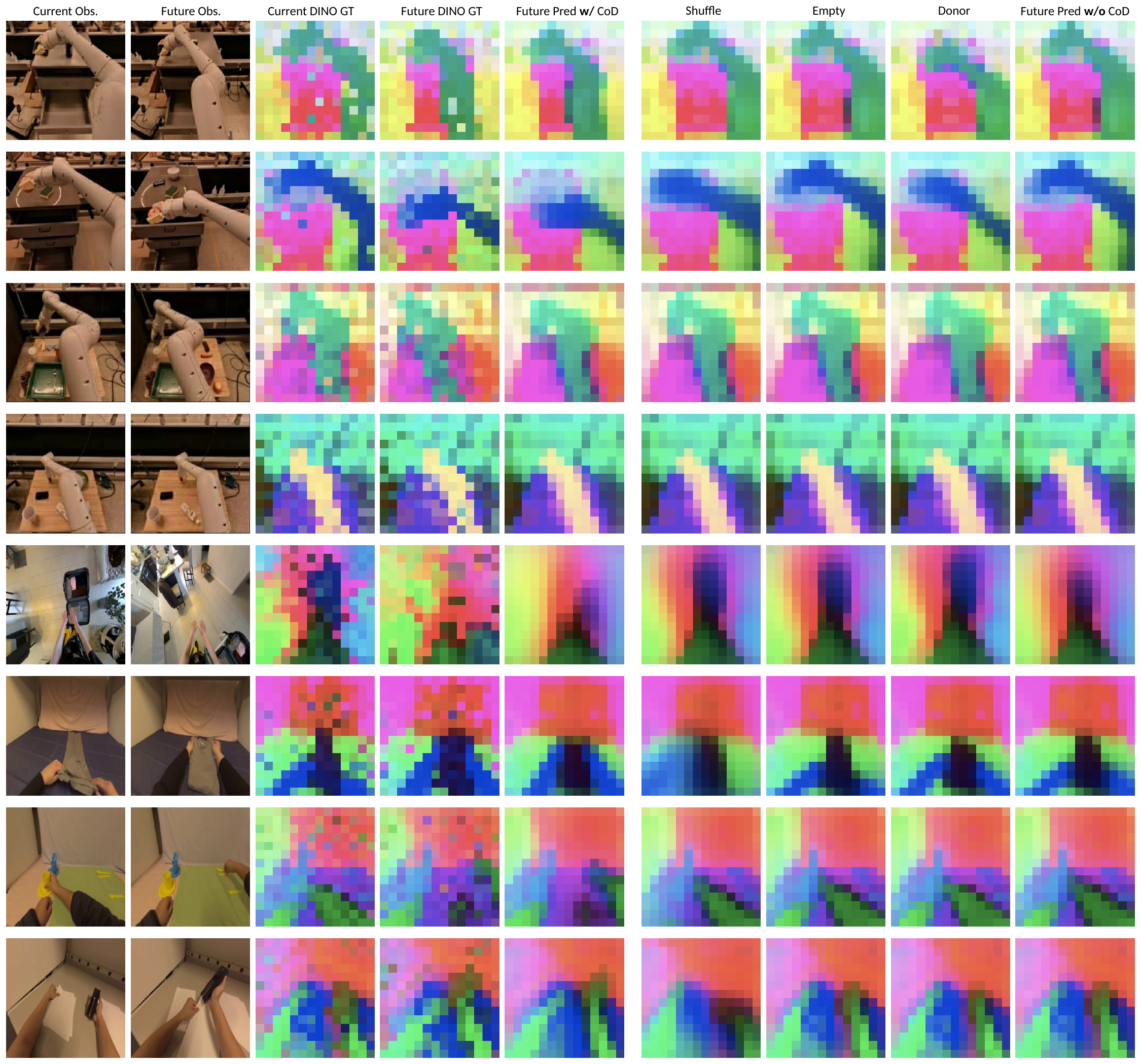}
  \caption{%
    \textbf{Qualitative future-prediction results.}
    Each row is one episode.
    The first two columns show the current and ground-truth future RGB
    observations, and the next two show PCA projections of the
    ground-truth DINO features for those frames.
    The remaining five columns show the predicted future DINO features
    under different CoD conditions: with CoD, shuffled CoD, empty CoD,
    a CoD taken from a different (donor) episode, and without CoD.
    All PCA panels share one color mapping. 
  }
\end{figure*}

\clearpage
\setcitestyle{numbers}
\bibliographystyle{plainnat}
\bibliography{main}

@String(CVPR= {IEEE Conf. Comput. Vis. Pattern Recog.})

@String(ICCV= {Int. Conf. Comput. Vis.})

@String(CVPR  = {CVPR})

@String(ICCV  = {ICCV})

@article{zheng2026egoscale,
  title={Egoscale: Scaling dexterous manipulation with diverse egocentric human data},
  author={Zheng, Ruijie and Niu, Dantong and Xie, Yuqi and Wang, Jing and Xu, Mengda and Jiang, Yunfan and Casta{\~n}eda, Fernando and Hu, Fengyuan and Tan, You Liang and Fu, Letian and others},
  journal={arXiv preprint arXiv:2602.16710},
  year={2026}
}

@inproceedings{li2026egowam,
  title={EgoWAM: World Action Models Beyond Pixels with In-the-Wild Egocentric Human Data},
  author={Li, Baoyu and Yin, Xinchen and Lin, Mengying and Zhang, Yixin and Xu, Danfei},
  booktitle={Robot World Models},
  year={2026}
}

@article{yang2025egovla,
  title={Egovla: Learning vision-language-action models from egocentric human videos},
  author={Yang, Ruihan and Yu, Qinxi and Wu, Yecheng and Yan, Rui and Li, Borui and Cheng, An-Chieh and Zou, Xueyan and Fang, Yunhao and Cheng, Xuxin and Qiu, Ri-Zhao and others},
  journal={arXiv preprint arXiv:2507.12440},
  year={2025}
}

@article{kim2026ego,
  title={Ego-Pi: VLA Fine-Tuning for Ego-Centric Human and Robot Data},
  author={Kim, Ji Woong and Wang, Ke and Fu, Zipeng and Chen, Sirui and Zhao, Cong and Lai, Jeff and Finn, Chelsea},
  journal={arXiv preprint arXiv:2606.08107},
  year={2026}
}

@inproceedings{chen2025moto,
  title={Moto: Latent motion token as the bridging language for learning robot manipulation from videos},
  author={Chen, Yi and Ge, Yuying and Tang, Weiliang and Li, Yizhuo and Ge, Yixiao and Ding, Mingyu and Shan, Ying and Liu, Xihui},
  booktitle={2025 IEEE/CVF International Conference on Computer Vision (ICCV)},
  pages={19752--19763},
  year={2025},
  organization={IEEE}
}

@inproceedings{routray2026vipra,
  title={Vipra: Video prediction for robot actions},
  author={Routray, Sandeep Kumar and Pan, Hengkai and Jain, Unnat and Bahl, Shikhar and Pathak, Deepak},
  booktitle={International Conference on Learning Representations},
  volume={2026},
  pages={68823--68857},
  year={2026}
}

@article{brohan2023rt,
  title={Rt-2: Vision-language-action models transfer web knowledge to robotic control},
  author={Brohan, Anthony and Brown, Noah and Carbajal, Justice and Chebotar, Yevgen and Chen, Xi and Choromanski, Krzysztof and Ding, Tianli and Driess, Danny and Dubey, Avinava and Finn, Chelsea and others},
  journal={arXiv preprint arXiv:2307.15818},
  year={2023}
}

@article{team2024octo,
  title={Octo: An open-source generalist robot policy},
  author={Team, Octo Model and Ghosh, Dibya and Walke, Homer and Pertsch, Karl and Black, Kevin and Mees, Oier and Dasari, Sudeep and Hejna, Joey and Kreiman, Tobias and Xu, Charles and others},
  journal={arXiv preprint arXiv:2405.12213},
  year={2024}
}

@article{kim24openvla,
    title={OpenVLA: An Open-Source Vision-Language-Action Model},
    author={{Moo Jin} Kim and Karl Pertsch and Siddharth Karamcheti and Ted Xiao and Ashwin Balakrishna and Suraj Nair and Rafael Rafailov and Ethan Foster and Grace Lam and Pannag Sanketi and Quan Vuong and Thomas Kollar and Benjamin Burchfiel and Russ Tedrake and Dorsa Sadigh and Sergey Levine and Percy Liang and Chelsea Finn},
    journal = {arXiv preprint arXiv:2406.09246},
    year={2024},
}

@inproceedings{o2024open,
  title={Open x-embodiment: Robotic learning datasets and rt-x models: Open x-embodiment collaboration 0},
  author={O’Neill, Abby and Rehman, Abdul and Maddukuri, Abhiram and Gupta, Abhishek and Padalkar, Abhishek and Lee, Abraham and Pooley, Acorn and Gupta, Agrim and Mandlekar, Ajay and Jain, Ajinkya and others},
  booktitle={2024 IEEE International Conference on Robotics and Automation (ICRA)},
  pages={6892--6903},
  year={2024},
  organization={IEEE}
}

@article{khazatsky2024droid,
  title={Droid: A large-scale in-the-wild robot manipulation dataset},
  author={Khazatsky, Alexander and Pertsch, Karl and Nair, Suraj and Balakrishna, Ashwin and Dasari, Sudeep and Karamcheti, Siddharth and Nasiriany, Soroush and Srirama, Mohan Kumar and Chen, Lawrence Yunliang and Ellis, Kirsty and others},
  journal={arXiv preprint arXiv:2403.12945},
  year={2024}
}

@article{wu2025robocoin,
  title={Robocoin: An open-sourced bimanual robotic data collection for integrated manipulation},
  author={Wu, Shihan and Liu, Xuecheng and Xie, Shaoxuan and Wang, Pengwei and Li, Xinghang and Yang, Bowen and Li, Zhe and Zhu, Kai and Wu, Hongyu and Liu, Yiheng and others},
  journal={arXiv preprint arXiv:2511.17441},
  year={2025}
}

@article{bjorck2025gr00t,
  title={Gr00t n1: An open foundation model for generalist humanoid robots},
  author={Bjorck, Johan and Casta{\~n}eda, Fernando and Cherniadev, Nikita and Da, Xingye and Ding, Runyu and Fan, Linxi and Fang, Yu and Fox, Dieter and Hu, Fengyuan and Huang, Spencer and others},
  journal={arXiv preprint arXiv:2503.14734},
  year={2025}
}

@article{black2024pi_0,
  title={$\pi_0$: A Vision-Language-Action Flow Model for General Robot Control},
  author={Black, Kevin and Brown, Noah and Driess, Danny and Esmail, Adnan and Equi, Michael and Finn, Chelsea and Fusai, Niccolo and Groom, Lachy and Hausman, Karol and Ichter, Brian and others},
  journal={arXiv preprint arXiv:2410.24164},
  year={2024}
}

@article{intelligence2025pi0,
  title={$\pi$0. 5: a vision-language-action model with open-world generalization, 2025},
  author={Intelligence, Physical and Black, Kevin and Brown, Noah and Darpinian, James and Dhabalia, Karan and Driess, Danny and Esmail, Adnan and Equi, Michael and Finn, Chelsea and Fusai, Niccolo and others},
  journal={URL https://arxiv. org/abs/2504.16054},
  volume={1},
  number={2},
  pages={3},
  year={2025}
}

@article{chi2023diffusion,
  title={Diffusion policy: Visuomotor policy learning via action diffusion},
  author={Chi, Cheng and Xu, Zhenjia and Feng, Siyuan and Cousineau, Eric and Du, Yilun and Burchfiel, Benjamin and Tedrake, Russ and Song, Shuran},
  journal={The International Journal of Robotics Research},
  pages={02783649241273668},
  year={2023},
  publisher={SAGE Publications Sage UK: London, England}
}

@article{wang2026qwen,
  title={Qwen-vla: Unifying vision-language-action modeling across tasks, environments, and robot embodiments},
  author={Wang, Qiuyue and Li, Mingsheng and Guan, Jian and Ye, Jinhui and Xie, Sicheng and Liu, Yitao and Chen, Junhao and Liang, Zhixuan and Zhang, Jie and Hu, Xintong and others},
  journal={arXiv preprint arXiv:2605.30280},
  year={2026}
}

@article{chen2025internvla,
  title={Internvla-m1: A spatially guided vision-language-action framework for generalist robot policy},
  author={Chen, Xinyi and Chen, Yilun and Fu, Yanwei and Gao, Ning and Jia, Jiaya and Jin, Weiyang and Li, Hao and Mu, Yao and Pang, Jiangmiao and Qiao, Yu and others},
  journal={arXiv preprint arXiv:2510.13778},
  year={2025}
}

@article{team2025gemini,
  title={Gemini robotics: Bringing ai into the physical world},
  author={Team, Gemini Robotics and Abeyruwan, Saminda and Ainslie, Joshua and Alayrac, Jean-Baptiste and Arenas, Montserrat Gonzalez and Armstrong, Travis and Balakrishna, Ashwin and Baruch, Robert and Bauza, Maria and Blokzijl, Michiel and others},
  journal={arXiv preprint arXiv:2503.20020},
  year={2025}
}

@article{pertsch2025fast,
  title={Fast: Efficient action tokenization for vision-language-action models},
  author={Pertsch, Karl and Stachowicz, Kyle and Ichter, Brian and Driess, Danny and Nair, Suraj and Vuong, Quan and Mees, Oier and Finn, Chelsea and Levine, Sergey},
  journal={arXiv preprint arXiv:2501.09747},
  year={2025}
}

@article{huang2026thinkact,
  title={Thinkact: Vision-language-action reasoning via reinforced visual latent planning},
  author={Huang, Chi-Pin and Wu, Yueh-Hua and Chen, Min-Hung and Wang, Frank and Yang, Fu-En},
  journal={Advances in Neural Information Processing Systems},
  volume={38},
  pages={82782--82802},
  year={2026}
}

@article{huang2026fast,
  title={Fast-thinkact: Efficient vision-language-action reasoning via verbalizable latent planning},
  author={Huang, Chi-Pin and Man, Yunze and Yu, Zhiding and Chen, Min-Hung and Kautz, Jan and Wang, Yu-Chiang Frank and Yang, Fu-En},
  journal={arXiv preprint arXiv:2601.09708},
  year={2026}
}

@article{lee2025molmoact,
  title={Molmoact: Action reasoning models that can reason in space},
  author={Lee, Jason and Duan, Jiafei and Fang, Haoquan and Deng, Yuquan and Liu, Shuo and Li, Boyang and Fang, Bohan and Zhang, Jieyu and Wang, Yi Ru and Lee, Sangho and others},
  journal={arXiv preprint arXiv:2508.07917},
  year={2025}
}

@article{fang2026molmoact2,
  title={Molmoact2: Action reasoning models for real-world deployment},
  author={Fang, Haoquan and Duan, Jiafei and Clay, Donovan and Wang, Sam and Liu, Shuo and Huang, Weikai and Fan, Xiang and Tsai, Wei-Chuan and Chen, Shirui and Wang, Yi Ru and others},
  journal={arXiv preprint arXiv:2605.02881},
  year={2026}
}

@article{zawalski2024robotic,
  title={Robotic control via embodied chain-of-thought reasoning},
  author={Zawalski, Micha{\l} and Chen, William and Pertsch, Karl and Mees, Oier and Finn, Chelsea and Levine, Sergey},
  journal={arXiv preprint arXiv:2407.08693},
  year={2024}
}

@article{zhong2026acot,
  title={Acot-vla: Action chain-of-thought for vision-language-action models},
  author={Zhong, Linqing and Liu, Yi and Wei, Yifei and Xiong, Ziyu and Yao, Maoqing and Liu, Si and Ren, Guanghui},
  journal={arXiv preprint arXiv:2601.11404},
  year={2026}
}

@article{zhang2026dreamvla,
  title={Dreamvla: a vision-language-action model dreamed with comprehensive world knowledge},
  author={Zhang, Wenyao and Liu, Hongsi and Qi, Zekun and Wang, Yunnan and Yu, Xinqiang and Zhang, Jiazhao and Dong, Runpei and He, Jiawei and Wang, He and Zhang, Zhizheng and others},
  journal={Advances in Neural Information Processing Systems},
  volume={38},
  pages={24195--24228},
  year={2026}
}

@article{li2026fomovla,
  title={FoMoVLA: Bridging Visual Foresight and Motion Guidance for Vision-Language-Action Models},
  author={Li, Wei and Jia, Peijin and Ma, Yuan and Jiang, Xuefeng and Jiang, Titong and Sun, Sheng and Li, Yujian and Wen, Xin and Hong, Han and Liu, Zhikang and others},
  journal={arXiv preprint arXiv:2607.14739},
  year={2026}
}

@article{gao2026dreamdojo,
  title={Dreamdojo: A generalist robot world model from large-scale human videos},
  author={Gao, Shenyuan and Liang, William and Zheng, Kaiyuan and Malik, Ayaan and Ye, Seonghyeon and Yu, Sihyun and Tseng, Wei-Cheng and Dong, Yuzhu and Mo, Kaichun and Lin, Chen-Hsuan and others},
  journal={arXiv preprint arXiv:2602.06949},
  year={2026}
}

@article{agarwal2026cosmos,
  title={Cosmos 3: Omnimodal world models for physical ai},
  author={Agarwal, Niket and Ali, Arslan and Allen, Jon and Antolini, Martin and Aubame, Adeline and Azzolini, Alisson and Bai, Junjie and Bala, Maciej and Balaji, Yogesh and Bapst, Josh and others},
  journal={arXiv preprint arXiv:2606.02800},
  year={2026}
}

@article{cai2026internvla,
  title={Internvla-a1: Unifying understanding, generation and action for robotic manipulation},
  author={Cai, Junhao and Cai, Zetao and Cao, Jiafei and Chen, Yilun and He, Zeyu and Jiang, Lei and Li, Hang and Li, Hengjie and Li, Yang and Liu, Yufei and others},
  journal={arXiv preprint arXiv:2601.02456},
  year={2026}
}

@article{ye2026world,
  title={World action models are zero-shot policies},
  author={Ye, Seonghyeon and Ge, Yunhao and Zheng, Kaiyuan and Gao, Shenyuan and Yu, Sihyun and Kurian, George and Indupuru, Suneel and Tan, You Liang and Zhu, Chuning and Xiang, Jiannan and others},
  journal={arXiv preprint arXiv:2602.15922},
  year={2026}
}

@article{pai2025mimic,
  title={mimic-video: Video-action models for generalizable robot control beyond vlas},
  author={Pai, Jonas and Achenbach, Liam and Montesinos, Victoriano and Forrai, Benedek and Mees, Oier and Nava, Elvis},
  journal={arXiv preprint arXiv:2512.15692},
  year={2025}
}

@article{hafner2025training,
  title={Training agents inside of scalable world models},
  author={Hafner, Danijar and Yan, Wilson and Lillicrap, Timothy},
  journal={arXiv preprint arXiv:2509.24527},
  year={2025}
}

@article{hafner2023mastering,
  title={Mastering diverse domains through world models},
  author={Hafner, Danijar and Pasukonis, Jurgis and Ba, Jimmy and Lillicrap, Timothy},
  journal={arXiv preprint arXiv:2301.04104},
  year={2023}
}

@article{zhou2024dino,
  title={Dino-wm: World models on pre-trained visual features enable zero-shot planning},
  author={Zhou, Gaoyue and Pan, Hengkai and LeCun, Yann and Pinto, Lerrel},
  journal={arXiv preprint arXiv:2411.04983},
  year={2024}
}

@article{assran2025vjepa2,
  title={V-JEPA~2: Self-Supervised Video Models Enable Understanding, Prediction and Planning},
  author={Assran, Mahmoud and Bardes, Adrien and Fan, David and Garrido, Quentin and Howes, Russell and
Komeili, Mojtaba and Muckley, Matthew and Rizvi, Ammar and Roberts, Claire and Sinha, Koustuv and Zholus, Artem and
Arnaud, Sergio and Gejji, Abha and Martin, Ada and Robert Hogan, Francois and Dugas, Daniel and
Bojanowski, Piotr and Khalidov, Vasil and Labatut, Patrick and Massa, Francisco and Szafraniec, Marc and
Krishnakumar, Kapil and Li, Yong and Ma, Xiaodong and Chandar, Sarath and Meier, Franziska and LeCun, Yann and
Rabbat, Michael and Ballas, Nicolas},
  journal={arXiv preprint arXiv:2506.09985},
  year={2025}
}

@article{kim2026cosmos,
  title={Cosmos policy: Fine-tuning video models for visuomotor control and planning},
  author={Kim, Moo Jin and Gao, Yihuai and Lin, Tsung-Yi and Lin, Yen-Chen and Ge, Yunhao and Lam, Grace and Liang, Percy and Song, Shuran and Liu, Ming-Yu and Finn, Chelsea and others},
  journal={arXiv preprint arXiv:2601.16163},
  year={2026}
}

@article{shao2024deepseekmath,
  title={Deepseekmath: Pushing the limits of mathematical reasoning in open language models},
  author={Shao, Zhihong and Wang, Peiyi and Zhu, Qihao and Xu, Runxin and Song, Junxiao and Bi, Xiao and Zhang, Haowei and Zhang, Mingchuan and Li, YK and Wu, Y and others},
  journal={arXiv preprint arXiv:2402.03300},
  year={2024}
}

@inproceedings{ye2025latent,
  title={Latent action pretraining from videos},
  author={Ye, Seonghyeon and Jang, Joel and Jeon, Byeongguk and Joo, Se June and Yang, Jianwei and Peng, Baolin and Mandlekar, Ajay and Tan, Reuben and Chao, Yu-Wei and Lin, Bill Yuchen and others},
  booktitle={International Conference on Learning Representations},
  volume={2025},
  pages={28213--28239},
  year={2025}
}

@article{bu2025univla,
  title={Univla: Learning to act anywhere with task-centric latent actions},
  author={Bu, Qingwen and Yang, Yanting and Cai, Jisong and Gao, Shenyuan and Ren, Guanghui and Yao, Maoqing and Luo, Ping and Li, Hongyang},
  journal={arXiv preprint arXiv:2505.06111},
  year={2025}
}

@article{dai2026conla,
  title={Conla: Contrastive latent action learning from human videos for robotic manipulation},
  author={Dai, Weisheng and Lan, Kai and Zhou, Jianyi and Zhao, Bo and Su, Xiu and Tong, Junwen and Guan, Weili and Yang, Shuo},
  journal={arXiv preprint arXiv:2602.00557},
  year={2026}
}

@inproceedings{sun2026vla,
  title={Vla-jepa: Enhancing vision-language-action model with latent world model},
  author={Sun, Jingwen and Zhang, Wenyao and Qi, Zekun and Ren, Shaojie and Liu, Zezhi and Zhu, Hanxin and Sun, Guangzhong and Jin, Xin and Chen, Zhibo},
  booktitle={European Conference on Computer Vision},
  pages={478--497},
  year={2026},
  organization={Springer}
}

@inproceedings{kareer2025egomimic,
  title={Egomimic: Scaling imitation learning via egocentric video},
  author={Kareer, Simar and Patel, Dhruv and Punamiya, Ryan and Mathur, Pranay and Cheng, Shuo and Wang, Chen and Hoffman, Judy and Xu, Danfei},
  booktitle={2025 IEEE International Conference on Robotics and Automation (ICRA)},
  pages={13226--13233},
  year={2025},
  organization={IEEE}
}

@article{liu2026egoengine,
  title={EgoEngine: From Egocentric Human Videos to High-Fidelity Dexterous Robot Demonstrations},
  author={Liu, Yangcen and Cheng, Shuo and Yin, Xinchen and Shin, Woo Chul and Cueva, Alfred and Yang, Yiran and Chen, Zhenyang and Zhang, Chuye and Xu, Danfei},
  journal={arXiv preprint arXiv:2606.12604},
  year={2026}
}

@article{du2023learning,
  title={Learning universal policies via text-guided video generation},
  author={Du, Yilun and Yang, Sherry and Dai, Bo and Dai, Hanjun and Nachum, Ofir and Tenenbaum, Josh and Schuurmans, Dale and Abbeel, Pieter},
  journal={Advances in neural information processing systems},
  volume={36},
  pages={9156--9172},
  year={2023}
}

@inproceedings{black2024zero,
  title={Zero-shot robotic manipulation with pre-trained image-editing diffusion models},
  author={Black, Kevin and Nakamoto, Mitsuhiko and Atreya, Pranav and Walke, Homer and Finn, Chelsea and Kumar, Aviral and Levine, Sergey},
  booktitle={International Conference on Learning Representations},
  volume={2024},
  pages={33431--33452},
  year={2024}
}

@article{bharadhwaj2024gen2act,
  title={Gen2act: Human video generation in novel scenarios enables generalizable robot manipulation},
  author={Bharadhwaj, Homanga and Dwibedi, Debidatta and Gupta, Abhinav and Tulsiani, Shubham and Doersch, Carl and Xiao, Ted and Shah, Dhruv and Xia, Fei and Sadigh, Dorsa and Kirmani, Sean},
  journal={arXiv preprint arXiv:2409.16283},
  year={2024}
}

@inproceedings{zhao2025cot,
  title={Cot-vla: Visual chain-of-thought reasoning for vision-language-action models},
  author={Zhao, Qingqing and Lu, Yao and Kim, Moo Jin and Fu, Zipeng and Zhang, Zhuoyang and Wu, Yecheng and Li, Zhaoshuo and Ma, Qianli and Han, Song and Finn, Chelsea and others},
  booktitle={2025 IEEE/CVF Conference on Computer Vision and Pattern Recognition (CVPR)},
  pages={1702--1713},
  year={2025},
  organization={IEEE}
}

@article{cen2025worldvla,
  title={Worldvla: Towards autoregressive action world model},
  author={Cen, Jun and Yu, Chaohui and Yuan, Hangjie and Jiang, Yuming and Huang, Siteng and Guo, Jiayan and Li, Xin and Song, Yibing and Luo, Hao and Wang, Fan and others},
  journal={arXiv preprint arXiv:2506.21539},
  year={2025}
}

@article{simeoni2025dinov3,
  title={Dinov3},
  author={Sim{\'e}oni, Oriane and Vo, Huy V and Seitzer, Maximilian and Baldassarre, Federico and Oquab, Maxime and Jose, Cijo and Khalidov, Vasil and Szafraniec, Marc and Yi, Seungeun and Ramamonjisoa, Micha{\"e}l and others},
  journal={arXiv preprint arXiv:2508.10104},
  year={2025}
}

@article{chen2026lawam,
  title={Lawam: Latent world action models for efficient dynamics-aware robot policies},
  author={Chen, Jialei and Wang, Kai and Chen, Kang and Chen, Shuaihang and Gao, Feng and Tang, Wenhao and Li, Zhiyuan and Liu, Weilin and Yao, Zhuyu and Li, Boxun and others},
  journal={arXiv preprint arXiv:2606.15768},
  year={2026}
}

@article{chen25ecotlite,
        title={Training Strategies for Efficient Embodied Reasoning},
        author={William Chen and Suneel Belkhale and Suvir Mirchandani and Oier Mees and Danny Driess and Karl Pertsch and Sergey Levine},
        journal = {arXiv preprint arXiv:2505.08243},
        year={2025},
}

@article{sun2026revisiting,
  title={Revisiting embodied chain-of-thought for generalizable robot manipulation},
  author={Sun, Nan and Zhang, Yuan and Yang, Yongkun and Zhao, Wentao and Li, Peiyan and Guo, Jun and Song, Wenxuan and Ding, Pengxiang and Suo, Runze and Su, Yifei and others},
  journal={arXiv preprint arXiv:2606.03784},
  year={2026}
}

@article{shah2026mv2mae,
  title={MV2MAE: Self-Supervised Video Pre-Training with Motion-Aware Multi-View Masked Autoencoders},
  author={Shah, Ketul and Crandall, Robert and Xu, Jie and Zhou, Peng and Pillai, Vipin and George, Marian and Bansal, Mayank and Chellappa, Rama},
  journal={Transactions on Machine Learning Research},
  year={2026}
}

@article{liu2026gdpo,
  title={Gdpo: Group reward-decoupled normalization policy optimization for multi-reward rl optimization},
  author={Liu, Shih-Yang and Dong, Xin and Lu, Ximing and Diao, Shizhe and Belcak, Peter and Liu, Mingjie and Chen, Min-Hung and Yin, Hongxu and Wang, Yu-Chiang Frank and Cheng, Kwang-Ting and others},
  journal={arXiv preprint arXiv:2601.05242},
  year={2026}
}

@article{bai2025qwen3,
  title={Qwen3-vl technical report},
  author={Bai, Shuai and Cai, Yuxuan and Chen, Ruizhe and Chen, Keqin and Chen, Xionghui and Cheng, Zesen and Deng, Lianghao and Ding, Wei and Gao, Chang and Ge, Chunjiang and others},
  journal={arXiv preprint arXiv:2511.21631},
  year={2025}
}

@article{driess2026knowledge,
  title={Knowledge insulating vision-language-action models: Train fast, run fast, generalize better},
  author={Driess, Danny and Springenberg, Jost and Ichter, Brian and Yu, Lili and Li-Bell, Adrian and Pertsch, Karl and Ren, Allen and Walke, Homer and Vuong, Quan and Shi, Lucy Xiaoyang and others},
  journal={Advances in Neural Information Processing Systems},
  volume={38},
  pages={102867--102888},
  year={2026}
}

@inproceedings{hoque2026egodex,
  title={Egodex: Learning dexterous manipulation from large-scale egocentric video},
  author={Hoque, Ryan and Huang, Peide and Yoon, David and Zhang, Jian and others},
  booktitle={International Conference on Learning Representations},
  volume={2026},
  pages={4218--4237},
  year={2026}
}

@article{punamiya2026egoverse,
  title={Egoverse: An egocentric human dataset for robot learning from around the world},
  author={Punamiya, Ryan and Kareer, Simar and Liu, Zeyi and Citron, Josh and Qiu, Ri-Zhao and Cai, Xiongyi and Gavryushin, Alexey and Chen, Jiaqi and Liconti, Davide and Zhu, Lawrence Y and others},
  journal={arXiv preprint arXiv:2604.07607},
  year={2026}
}

@article{zhou2025libero,
  title={Libero-pro: Towards robust and fair evaluation of vision-language-action models beyond memorization},
  author={Zhou, Xueyang and Xu, Yangming and Tie, Guiyao and Chen, Yongchao and Zhang, Guowen and Chu, Duanfeng and Zhou, Pan and Sun, Lichao},
  journal={arXiv preprint arXiv:2510.03827},
  year={2025}
}

@article{liu2023libero,
  title={Libero: Benchmarking knowledge transfer for lifelong robot learning},
  author={Liu, Bo and Zhu, Yifeng and Gao, Chongkai and Feng, Yihao and Liu, Qiang and Zhu, Yuke and Stone, Peter},
  journal={Advances in Neural Information Processing Systems},
  volume={36},
  pages={44776--44791},
  year={2023}
}

@inproceedings{walke2023bridgedata,
    title={BridgeData V2: A Dataset for Robot Learning at Scale},
    author={Walke, Homer and Black, Kevin and Lee, Abraham and Kim, Moo Jin and Du, Max and Zheng, Chongyi and Zhao, Tony and Hansen-Estruch, Philippe and Vuong, Quan and He, Andre and Myers, Vivek and Fang, Kuan and Finn, Chelsea and Levine, Sergey},
    booktitle={Conference on Robot Learning (CoRL)},
    year={2023}
}

@inproceedings{jang2021bc,
title={{BC}-Z: Zero-Shot Task Generalization with Robotic Imitation Learning},
author={Eric Jang and Alex Irpan and Mohi Khansari and Daniel Kappler and Frederik Ebert and Corey Lynch and Sergey Levine and Chelsea Finn},
booktitle={5th Annual Conference on Robot Learning},
year={2021},
url={https://openreview.net/forum?id=8kbp23tSGYv}}

@inproceedings{deitke2025molmo,
  title={Molmo and pixmo: Open weights and open data for state-of-the-art vision-language models},
  author={Deitke, Matt and Clark, Christopher and Lee, Sangho and Tripathi, Rohun and Yang, Yue and Park, Jae Sung and Salehi, Mohammadreza and Muennighoff, Niklas and Lo, Kyle and Soldaini, Luca and others},
  booktitle={2025 IEEE/CVF Conference on Computer Vision and Pattern Recognition (CVPR)},
  pages={91--104},
  year={2025},
  organization={IEEE}
}

@inproceedings{clark2026molmo2,
  title={Molmo2: Open weights and data for vision-language models with video understanding and grounding},
  author={Clark, Christopher and Zhang, Jieyu and Ma, Zixian and Park, Jae Sung and Tripathi, Rohun and Lee, Sangho and Salehi, Mohammadreza and Ren, Jason and Kim, Chris Dongjoo and Yang, Yinuo and others},
  booktitle={Proceedings of the IEEE/CVF Conference on Computer Vision and Pattern Recognition},
  pages={28652--28668},
  year={2026}
}

@InProceedings{robopoint,
  title = 	 {RoboPoint: A Vision-Language Model for Spatial Affordance Prediction in Robotics},
  author =       {Yuan, Wentao and Duan, Jiafei and Blukis, Valts and Pumacay, Wilbert and Krishna, Ranjay and Murali, Adithyavairavan and Mousavian, Arsalan and Fox, Dieter},
  booktitle = 	 {Proceedings of The 8th Conference on Robot Learning},
  pages = 	 {4005--4020},
  year = 	 {2025},
  editor = 	 {Agrawal, Pulkit and Kroemer, Oliver and Burgard, Wolfram},
  volume = 	 {270},
  series = 	 {Proceedings of Machine Learning Research},
  month = 	 {06--09 Nov},
  publisher =    {PMLR},
  url = 	 {https://proceedings.mlr.press/v270/yuan25c.html}
}

@inproceedings{robovqa,
    title={RoboVQA: Multimodal Long-Horizon Reasoning for Robotics},
    author={Pierre Sermanet and Tianli Ding and Jeffrey Zhao and Fei Xia and Debidatta Dwibedi and Keerthana Gopalakrishnan and Christine Chan and Gabriel Dulac-Arnold and Sharath Maddineni and Nikhil J Joshi and Pete Florence and Wei Han and Robert Baruch and Yao Lu and Suvir Mirchandani and Peng Xu and Pannag Sanketi and Karol Hausman and Izhak Shafran and Brian Ichter and Yuan Cao},
    booktitle={arXiv preprint arXiv:2311.00899},
    year={2023}
}

@misc{cadene2024lerobot,
    author = {Cadene, Remi and Alibert, Simon and Soare, Alexander and Gallouedec, Quentin and Zouitine, Adil and Palma, Steven and Kooijmans, Pepijn and Aractingi, Michel and Shukor, Mustafa and Aubakirova, Dana and Russi, Martino and Capuano, Francesco and Pascal, Caroline and Choghari, Jade and Meftah, Khalil and Ellerbach, Maxime and Moss, Jess and Wolf, Thomas},
    title = {LeRobot: State-of-the-art Machine Learning for Real-World Robotics in Pytorch},
    howpublished = "\url{https://github.com/huggingface/lerobot}",
    year = {2024}
}

@InProceedings{fei2026liberoplus,
    author    = {Fei, Senyu and Wang, Siyin and Shi, Junhao and Dai, Zihao and Cai, Jikun and Qian, Pengfang and Ji, Li and He, Xinzhe and Zhang, Shiduo and Fei, Zhaoye and Fu, Jinlan and Gong, Jingjing and Qiu, Xipeng},
    title     = {LIBERO-Plus: A Progressive Robustness Benchmark for Visual-Language-Action Models},
    booktitle = {Proceedings of the IEEE/CVF Conference on Computer Vision and Pattern Recognition (CVPR)},
    month     = {June},
    year      = {2026},
    pages     = {38574-38583}
}

@article{li2024evaluating,
  title={Evaluating real-world robot manipulation policies in simulation},
  author={Li, Xuanlin and Hsu, Kyle and Gu, Jiayuan and Pertsch, Karl and Mees, Oier and Walke, Homer Rich and Fu, Chuyuan and Lunawat, Ishikaa and Sieh, Isabel and Kirmani, Sean and others},
  journal={arXiv preprint arXiv:2405.05941},
  year={2024}
}

@article{li2025eagle,
  title={Eagle 2: Building post-training data strategies from scratch for frontier vision-language models},
  author={Li, Zhiqi and Chen, Guo and Liu, Shilong and Wang, Shihao and VS, Vibashan and Ji, Yishen and Lan, Shiyi and Zhang, Hao and Zhao, Yilin and Radhakrishnan, Subhashree and others},
  journal={arXiv preprint arXiv:2501.14818},
  year={2025}
}

@article{chen2026eagle,
  title={Eagle 2.5: Boosting long-context post-training for frontier vision-language models},
  author={Chen, Guo and Li, Zhiqi and Wang, Shihao and Jiang, Jindong and Liu, Yicheng and Lu, Lidong and Huang, De-An and Byeon, Wonmin and Le, Matthieu and Ehrlich, Max and others},
  journal={Advances in Neural Information Processing Systems},
  volume={38},
  pages={91077--91100},
  year={2026}
}

@inproceedings{ye2026omnivinci,
  title={Omnivinci: Enhancing architecture and data for omni-modal understanding llm},
  author={Ye, Hanrong and Yang, Chao-Han Huck and Goel, Arushi and Huang, Wei and Wan, Zhen and Tian, Jinchuan and Cheng, An-Chieh and Zhu, Ligeng and Su, Yuanhang and Lou, Yuming and others},
  booktitle={International Conference on Learning Representations},
  volume={2026},
  pages={56101--56138},
  year={2026}
}

@inproceedings{liu2025nvila,
  title={Nvila: Efficient frontier visual language models},
  author={Liu, Zhijian and Zhu, Ligeng and Shi, Baifeng and Zhang, Zhuoyang and Lou, Yuming and Yang, Shang and Xi, Haocheng and Cao, Shiyi and Gu, Yuxian and Li, Dacheng and others},
  booktitle={2025 IEEE/CVF Conference on Computer Vision and Pattern Recognition (CVPR)},
  pages={4122--4134},
  year={2025},
  organization={IEEE}
}

@article{lu2026tacitvla,
  title={TacitVLA: Reasoning Vision-Language-Action Models Without Reasoning Latency},
  author={Lu, Jack and You, Yurong and Li, Boyi and Cao, Yulong and Tian, Thomas and Xiao, Chaowei and Chen, Xiangyu and Luo, Wenjie and Ivanovic, Boris and Pavone, Marco and others},
  journal={arXiv preprint arXiv:2610.17540},
  year={2026}
}

@article{liu2026long,
  title={Long-horizon manipulation via trace-conditioned vla planning},
  author={Liu, Isabella and Cheng, An-Chieh and Yan, Rui and Chen, Geng and Qiu, Ri-Zhao and Zou, Xueyan and Yi, Sha and Yin, Hongxu and Wang, Xiaolong and Liu, Sifei},
  journal={arXiv preprint arXiv:2604.21924},
  year={2026}
}

@inproceedings{xu2023open,
  title={Open-vocabulary panoptic segmentation with text-to-image diffusion models},
  author={Xu, Jiarui and Liu, Sifei and Vahdat, Arash and Byeon, Wonmin and Wang, Xiaolong and De Mello, Shalini},
  booktitle={2023 IEEE/CVF conference on computer vision and pattern recognition (CVPR)},
  pages={2955--2966},
  year={2023},
  organization={IEEE}
}

@inproceedings{tosi2024diffusion,
  title={Diffusion models for monocular depth estimation: Overcoming challenging conditions},
  author={Tosi, Fabio and Ramirez, Pierluigi Zama and Poggi, Matteo},
  booktitle={European Conference on Computer Vision},
  pages={236--257},
  year={2024},
  organization={Springer}
}

@article{kim2025seg4diff,
  title={Seg4diff: Unveiling open-vocabulary segmentation in text-to-image diffusion transformers},
  author={Kim, Chaehyun and Shin, Heeseong and Hong, Eunbeen and Yoon, Heeji and Arnab, Anurag and Seo, Paul Hongsuck and Hong, Sunghwan and Kim, Seungryong},
  journal={arXiv preprint arXiv:2509.18096},
  year={2025}
}

@article{agarwal2025cosmos,
  title={Cosmos world foundation model platform for physical ai},
  author={Agarwal, Niket and Ali, Arslan and Bala, Maciej and Balaji, Yogesh and Barker, Erik and Cai, Tiffany and Chattopadhyay, Prithvijit and Chen, Yongxin and Cui, Yin and Ding, Yifan and others},
  journal={arXiv preprint arXiv:2501.03575},
  year={2025}
}

@article{wan2025wan,
  title={Wan: Open and advanced large-scale video generative models},
  author={Wan, Team and Wang, Ang and Ai, Baole and Wen, Bin and Mao, Chaojie and Xie, Chen-Wei and Chen, Di and Yu, Feiwu and Zhao, Haiming and Yang, Jianxiao and others},
  journal={arXiv preprint arXiv:2503.20314},
  year={2025}
}

@inproceedings{huang2025videomage,
  title={Videomage: Multi-subject and motion customization of text-to-video diffusion models},
  author={Huang, Chi-Pin and Wu, Yen-Siang and Chung, Hung-Kai and Chang, Kai-Po and Yang, Fu-En and Wang, Yu-Chiang Frank},
  booktitle={2025 IEEE/CVF Conference on Computer Vision and Pattern Recognition (CVPR)},
  pages={17603--17612},
  year={2025},
  organization={IEEE}
}

@article{li2026hydra,
  title={Hydra-0: Action Flow for Generalist World Modeling and Control},
  author={Li, Hongyu and Wen, Bowen and Zhu, Xinghao and Wang, Yixuan and Du, Yilun and Li, Yunzhu and Konidaris, George and Birchfield, Stan and Pouya, Soha and Li, Chenran and others},
  journal={arXiv preprint arXiv:2608.18077},
  year={2026}
}

@article{kim2026pri4r,
  title={Pri4r: Learning world dynamics for vision-language-action models with privileged 4d representation},
  author={Kim, Jisoo and Cho, Jungbin and Chu, Sanghyeok and Bal, Ananya and Kim, Jinhyung and Lee, Gunhee and Lee, Sihaeng and Kim, Seung Hwan and Han, Bohyung and Lee, Hyunmin and others},
  journal={arXiv preprint arXiv:2603.01549},
  year={2026}
}

\end{document}